\documentclass[10pt,journal,compsoc]{IEEEtran}
\usepackage{xcolor}
\usepackage{enumitem}
\usepackage{multirow}
\usepackage{booktabs}
\usepackage{makecell}
\makeatletter
\let\NAT@parse\undefined
\makeatother
\usepackage{hyperref}
\ifCLASSOPTIONcompsoc
  \usepackage[nocompress]{cite}
\else
  \usepackage{cite}
\fi
\ifCLASSINFOpdf
  \usepackage[pdftex]{graphicx}
  \usepackage{subfigure}
\else
\fi
\usepackage{amsmath}
\usepackage{amsfonts}

\usepackage{stfloats}
\begin{document}
%
% paper title
% Titles are generally capitalized except for words such as a, an, and, as,
% at, but, by, for, in, nor, of, on, or, the, to and up, which are usually
% not capitalized unless they are the first or last word of the title.
% Linebreaks \\ can be used within to get better formatting as desired.
% Do not put math or special symbols in the title.
\title{Distantly-Supervised Long-Tailed Relation Extraction Using Constraint Graphs}
%
%
% author names and IEEE memberships
% note positions of commas and nonbreaking spaces ( ~ ) LaTeX will not break
% a structure at a ~ so this keeps an author's name from being broken across
% two lines.
% use \thanks{} to gain access to the first footnote area
% a separate \thanks must be used for each paragraph as LaTeX2e's \thanks
% was not built to handle multiple paragraphs
%
%
%\IEEEcompsocitemizethanks is a special \thanks that produces the bulleted
% lists the Computer Society journals use for "first footnote" author
% affiliations. Use \IEEEcompsocthanksitem which works much like \item
% for each affiliation group. When not in compsoc mode,
% \IEEEcompsocitemizethanks becomes like \thanks and
% \IEEEcompsocthanksitem becomes a line break with idention. This
% facilitates dual compilation, although admittedly the differences in the
% desired content of \author between the different types of papers makes a
% one-size-fits-all approach a daunting prospect. For instance, compsoc 
% journal papers have the author affiliations above the "Manuscript
% received ..."  text while in non-compsoc journals this is reversed. Sigh.

\author{Tianming~Liang,
        Yang~Liu,
        Xiaoyan~Liu,
        Hao~Zhang,
        Gaurav~Sharma,~\IEEEmembership{Fellow,~IEEE}
        and~Maozu~Guo% <-this % stops a space
\IEEEcompsocitemizethanks{
  \IEEEcompsocthanksitem T.~Liang, Y.~Liu, X.~Liu, H.~Zhang are with the School of Computer Science and Technology, Harbin Institute of Technology,
  Harbin 150001, China. Email: {liangtianming, yliu76, liuxiaoyan}@hit.edu.cn and zhanghao2020@stu.hit.edu.cn.
  \IEEEcompsocthanksitem G. Sharma is with  the Department of
  Electrical and Computer Engineering, University of Rochester, Rochester,
  NY 14627, USA. Email: gaurav.sharma@rochester.edu.
  \IEEEcompsocthanksitem M. Guo is with the School of Electrical and Information Engineering,
  Beijing University of Civil Engineering and Architecture, Beijing 100044,
  China, and also with the Beijing Key Laboratory of Intelligent Processing
  for Building Big Data, Beijing University of Civil Engineering and Architecture, Beijing 100044, China. 
  Email: guomaozu@bucea.edu.cn.
  \IEEEcompsocthanksitem Corresponding author: Yang~Liu.
}}

% note the % following the last \IEEEmembership and also \thanks - 
% these prevent an unwanted space from occurring between the last author name
% and the end of the author line. i.e., if you had this:
% 
% \author{....lastname \thanks{...} \thanks{...} }
%                     ^------------^------------^----Do not want these spaces!
%
% a space would be appended to the last name and could cause every name on that
% line to be shifted left slightly. This is one of those "LaTeX things". For
% instance, "\textbf{A} \textbf{B}" will typeset as "A B" not "AB". To get
% "AB" then you have to do: "\textbf{A}\textbf{B}"
% \thanks is no different in this regard, so shield the last } of each \thanks
% that ends a line with a % and do not let a space in before the next \thanks.
% Spaces after \IEEEmembership other than the last one are OK (and needed) as
% you are supposed to have spaces between the names. For what it is worth,
% this is a minor point as most people would not even notice if the said evil
% space somehow managed to creep in.

% The paper headers
\markboth{IEEE TRANSACTIONS ON KNOWLEDGE AND DATA ENGINEERING}%
% \markboth{}%
{Shell \MakeLowercase{\textit{et al.}}: Bare Demo of IEEEtran.cls for Computer Society Journals}
% The only time the second header will appear is for the odd numbered pages
% after the title page when using the twoside option.
% 
% *** Note that you probably will NOT want to include the author's ***
% *** name in the headers of peer review papers.                   ***
% You can use \ifCLASSOPTIONpeerreview for conditional compilation here if
% you desire.

% The publisher's ID mark at the bottom of the page is less important with
% Computer Society journal papers as those publications place the marks
% outside of the main text columns and, therefore, unlike regular IEEE
% journals, the available text space is not reduced by their presence.
% If you want to put a publisher's ID mark on the page you can do it like
% this:
%\IEEEpubid{0000--0000/00\$00.00~\copyright~2015 IEEE}
% or like this to get the Computer Society new two part style.
%\IEEEpubid{\makebox[\columnwidth]{\hfill 0000--0000/00/\$00.00~\copyright~2015 IEEE}%
%\hspace{\columnsep}\makebox[\columnwidth]{Published by the IEEE Computer Society\hfill}}
% Remember, if you use this you must call \IEEEpubidadjcol in the second
% column for its text to clear the IEEEpubid mark (Computer Society jorunal
% papers don't need this extra clearance.)

\IEEEtitleabstractindextext{%
\begin{abstract}
  Label noise and long-tailed distributions are two major challenges in distantly supervised relation extraction.
  Recent studies have shown great progress on denoising, but paid little attention to the problem of long-tailed relations.
  In this paper, we introduce a constraint graph to model the dependencies between relation labels.
  On top of that, 
  we further propose a novel constraint graph-based relation extraction framework(CGRE)
  to handle the two challenges simultaneously.
  CGRE employs graph convolution networks to propagate information from data-rich relation nodes to data-poor relation nodes,
  and thus boosts the representation learning of long-tailed relations.
  To further improve the noise immunity, a constraint-aware attention module is designed in CGRE to integrate the constraint information.
  Extensive experimental results indicate that CGRE achieves significant improvements over the previous methods for both denoising and long-tailed relation extraction.

\end{abstract}

% Note that keywords are not normally used for peerreview papers.
}

% make the title area
\maketitle

% To allow for easy dual compilation without having to reenter the
% abstract/keywords data, the \IEEEtitleabstractindextext text will
% not be used in maketitle, but will appear (i.e., to be "transported")
% here as \IEEEdisplaynontitleabstractindextext when the compsoc 
% or transmag modes are not selected <OR> if conference mode is selected 
% - because all conference papers position the abstract like regular
% papers do.
\IEEEdisplaynontitleabstractindextext
% \IEEEdisplaynontitleabstractindextext has no effect when using
% compsoc or transmag under a non-conference mode.

% For peer review papers, you can put extra information on the cover
% page as needed:
% \ifCLASSOPTIONpeerreview
% \begin{center} \bfseries EDICS Category: 3-BBND \end{center}
% \fi
%
% For peerreview papers, this IEEEtran command inserts a page break and
% creates the second title. It will be ignored for other modes.
\IEEEpeerreviewmaketitle

\IEEEraisesectionheading{\section{Introduction}\label{sec:introduction}}
\IEEEPARstart{R}{elation} extraction (RE), which aims to extract the semantic relations between two entities from unstructured text, 
    is crucial for many natural language processing applications,
    such as knowledge graph completion~\cite{shen2020modeling,xiao2021knowledge}, 
    search engines~\cite{jiang2016generating, kalashnikov2008web}
    and question answering~\cite{hu2018answering, hua2020few}.
    Although conventional supervised approaches have been extensively researched, 
    they are still limited by the scarcity of manually annotated data. 
    Distantly supervised relation extraction (DSRE)~\cite{mintz2009distant} is one of the most promising techniques to address this problem, 
    because it can automatically generate large scale labeled data by aligning the entity pairs between text and knowledge bases (KBs).
    However, distant supervision suffers from two major challenges when used for RE.

    The first challenge in DSRE is \textbf{label noise}, which is caused by the distant supervision assumption:
    if one entity pair has a relationship in existing KBs, then all sentences mentioning the entity pair express this relation.
    For example, due to the relational triple (\textit{Bill Gates}, \textit{Founded}, \textit{Microsoft}), distant supervision will generate a noisy label \textit{Founded} for the sentence "\textit{Bill Gates} speaks at a conference held by \textit{Microsoft}",
    although this sentence does not mention this relation at all. 
    
    In recent years, many efforts have been devoted to improving the robustness of RE models against label noise.
    The combination of multi-instance learning and attention mechanism is one of the most popular strategies 
    to reduce the influence of label noise~\cite{lin2016neural, yuan2019distant, ye2019distant}.
    This strategy extracts relations of entity pairs from sentence bags,
    with the objective of alleviating the sentence-level label noise.
    In addition, some novel strategies, 
    such as reinforcement learning~\cite{jun2018reinforcement, li2020adaptive}, 
    adversarial training~\cite{wu2017adversarial, qin2018dsgan, li2019gan, puspitaningrum2019improving}
    and deep clustering~\cite{zhang2017noise, shang2020noisy}
    also show great potential for DSRE.
    However, these approaches are driven totally by the noisy labeling data,
    which may misguide the optimization of parameters and further hurt the reliability of models.

    To address this problem, some researchers attempt to enrich the background knowledge of models
    by integrating external information. 
    In general, the external information, 
    e.g., entity descriptions~\cite{ji2017distant, hu2019improving}, entity types~\cite{liu2014exploring, vashishth2018reside, kuang2020improving} 
    and knowledge graphs~\cite{wang2018label, hu2019improving},
    will be encoded as vector form, and then integrated to DSRE models by simple concatenation or attention mechanism.
    Compared with the above implicit knowledge, constraint rules are explicit and direct information that can effectively enhance the discernment of models in noisy instances.
    For example, the relation for the sentence "\textit{Bill Gates} was 19 when he and Paul Allen started \textit{Microsoft}" can not be \textit{Child\_of}, since its head/tail entity must be a \textit{Person}, while \textit{Microsoft} is an \textit{Organization}. 
    Here, we define (\textit{Person}, \textit{Child\_of}, \textit{Person}) as a constraint for \textit{Child\_of}.
    However, directly removing the constraint-violating sentences from a dataset would result in loss of significant useful information (as demonstrated in Sec. \ref{sec3.6.2}).
    Hence, we explore a soft way to integrate the constraint information by attention mechanism in this paper. 

    \begin{figure}[t]
      \centering
      \includegraphics[scale=0.6]{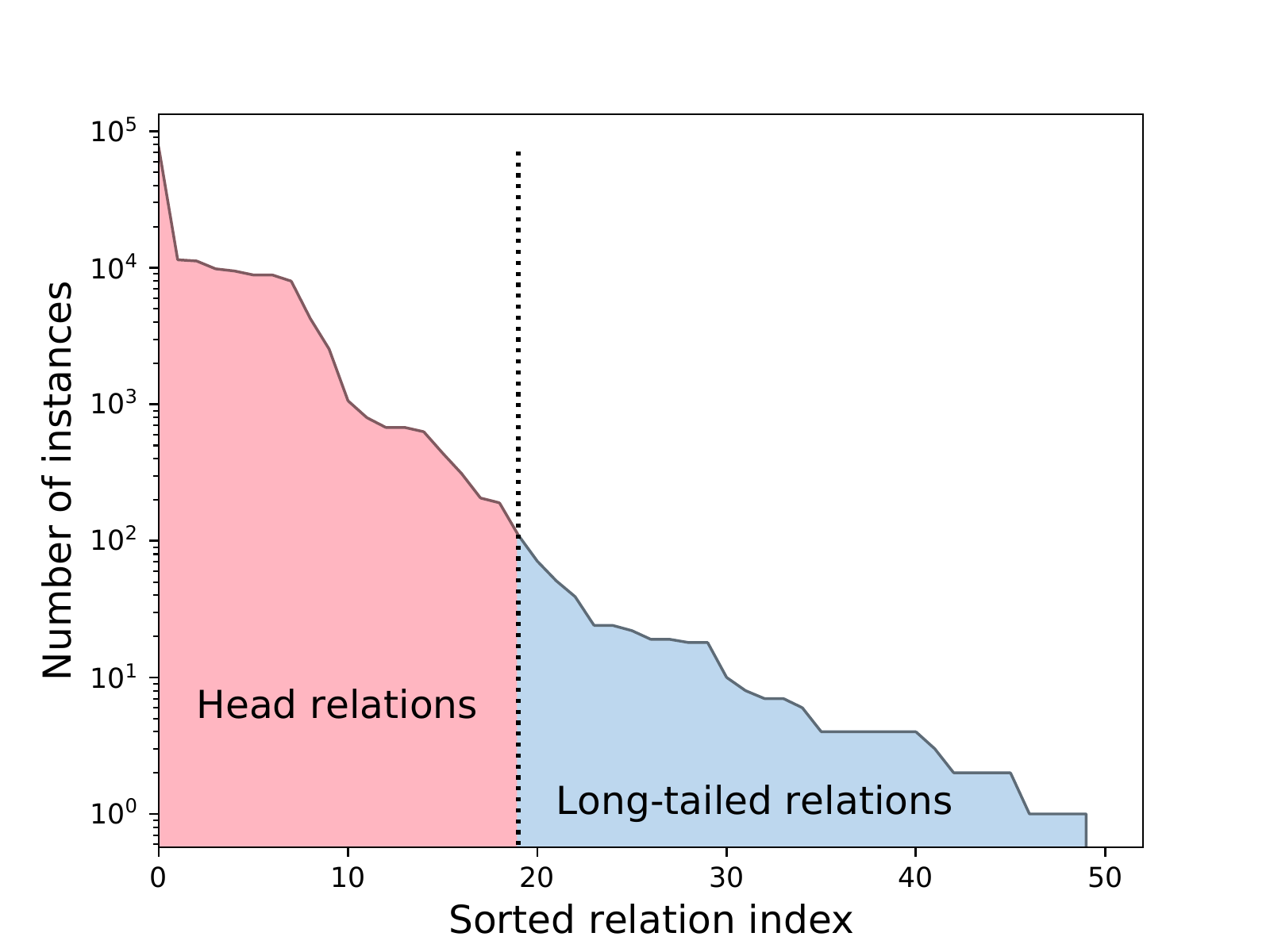}
      \caption{Instance number distribution of 52 positive relations in FB-NYT dataset. 
      Relations with more than 100 instances are allocated to \textit{head relation}, 
      while the remaining relations belong to \textit{long-tailed relation}.
      Note that the vertical axis uses a logarithmic scale.}
      \label{figure1}
    \end{figure}

    The second challenge in DSRE is \textbf{long-tailed relation extraction}, 
    however, which tends to be neglected as compared with the noisy labeling problem.
    In fact, real-world datasets of distant supervision always have a skewed distribution with a long tail, 
    i.e., a small proportion of relations (a.k.a \textit{head relation}) occupy most of the data, while most relations (a.k.a \textit{long-tailed relation}) merely have a few training instances.
    As shown in Figure \ref{figure1}, more than 60\% relations are long-tailed with fewer than 100 instances in the popular New York Times (FB-NYT) dataset~\cite{riedel2010modeling}.
    Long-tailed relation extraction is important for knowledge graph construction and completion.
    Unfortunately, even the state-of-the-art DSRE models are not able to handle long-tailed relations well.
    Hence, how to train a balanced relation extractor from unbalanced data becomes a serious problem in relation extraction.

    In general, long-tailed classfication tasks are addressed by re-sampling or re-weighting,
    nevertheless, long-tailed RE is special
    because the relation labels are semantically related rather than independent of each other.
    For example, if relation \textit{/location/country/capital} holds, 
    another relation \textit{/location/location/contains} will holds as well.
    Therefore, a promising strategy for long-tailed RE is 
    to mine latent relationships between relations by modeling the dependency paths.
    Once the dependency paths have been constructed, 
    the relation labels are no longer independent of each other, 
    and rich information can be propagated among the relations through these paths,
    which is crucial for the data-poor long-tailed relations. 
    To accomplish this, \cite{han2018hierarchical} proposed the relation hierarchical tree, 
    which connect different relation labels according to hierarchical information in the relation names.
    For instance, relation \textit{/people/person/nationality} and relation \textit{/people/person/religion} 
    have the same parent node \textit{/people/person} in the relation hierarchical tree.
    On account of its effectiveness, most of the 
    previous long-tailed RE models~\cite{han2018hierarchical, zhang2019long,  cao2021learning, li2020improving}
    rely on the relation hierarchical tree.
    However, relation hierarchical trees suffer from several limitations:

    (1) the construction of the relation hierarchical trees requires the relation names in hierarchical format, 
    which conflicts with many existing RE datasets, 
    such as SemEval-2010 Task8~\cite{hendrickx2009semeval}, TACRED~\cite{zhang2017position}, and FewRel~\cite{han2018fewrel, gao2019fewrel};
    
    (2) the sparsity of the hierarchy tree hinders the representation learning of some extremely long-tailed relations;
    
    (3) the optimization of parameters is purely driven by the noisy training data, lacking enough effective supervision signals. 
    Specifically, the key goal of selective attention mechanism is to automatically recognize the noisy sentences and then reduce their weights, but this is not guaranteed.
    Although pre-trained knowledge graph embeddings were utilized by \cite{zhang2019long} to guide the learning of attention mechanism, they were simply used as initial embeddings of the relation labels.

\begin{figure}[t]
      \centering
      \includegraphics[scale=0.4]{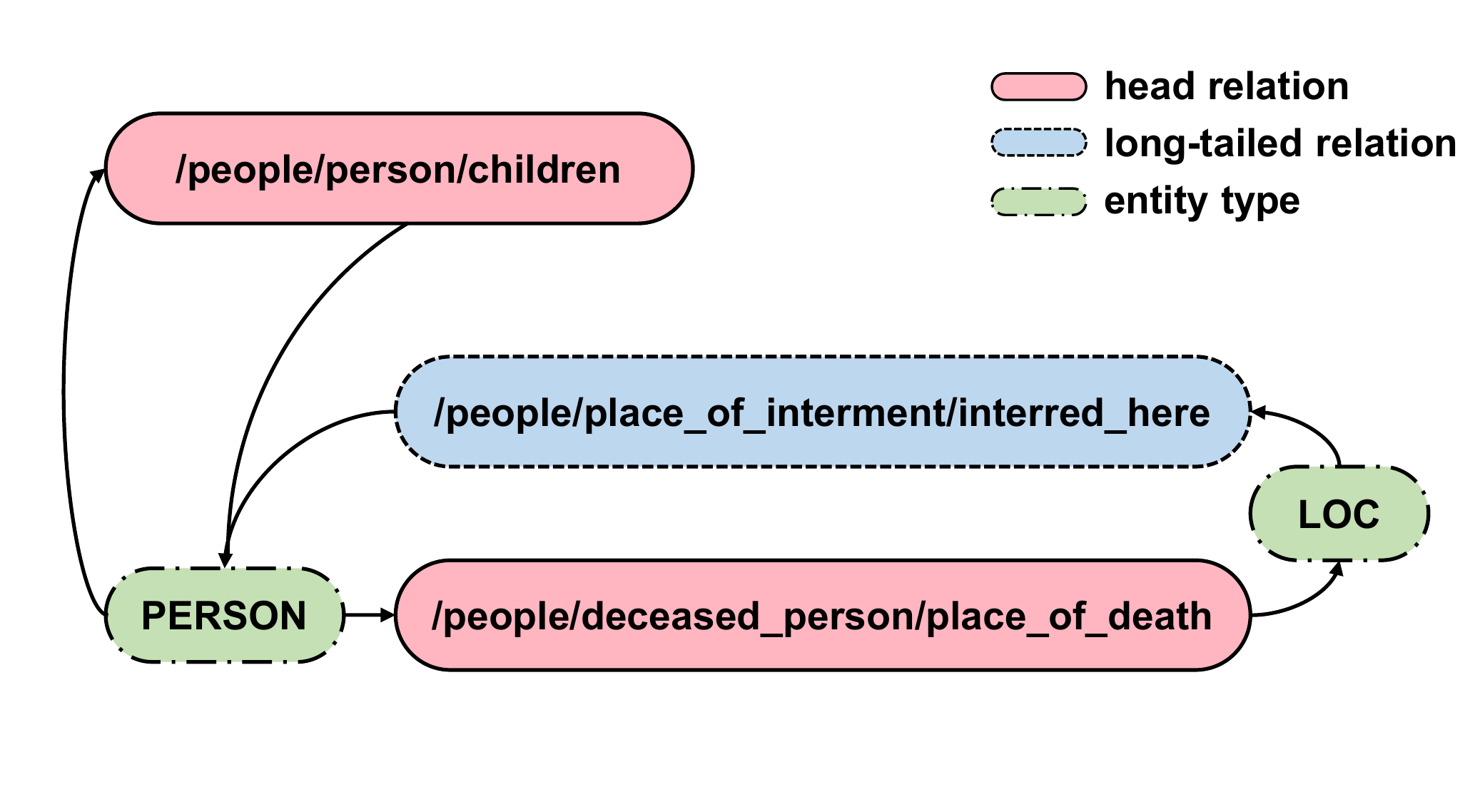}
      \caption{A subgraph of the constraint graph for FB-NYT.
      In this graph, the types of head and tail entities are respectively predecessors and successors of the corresponding relations.
      }
      \label{figure2}
    \end{figure}

    To overcome the limitations of relation hierarchies, 
    we explore the feasibility of utilizing another structure, constraint graph,
    to represent the intrinsic connection between relations.
    As shown in Figure \ref{figure2}, a constraint graph is a bipartite digraph~\cite{bang2008digraphs}, in which each directed edge connects a relation node to a type node or vice versa.
    As compared with entity-level knowledge graphs, 
    constraint graphs provide more basic and general knowledge, more direct rule constraints, and a smaller vector space for knowledge mining and reasoning.
    Moreover, most knowledge bases (e.g. Freebase) preserve the entity type constraint information~\cite{lei2018cooperative}, which can be directly used for constraint graph construction.
    Even for datasets without relevant information, the constraint graph can be easily constructed according to the definition of each relation or the co-occurrence frequency of each relation with entity type pairs in the training set.
    
    In our observations, there are at least three kinds of information in constraint graphs that are beneficial to DSRE:
    (1) \textbf{type information.} 
    As shown in many previous works \cite{vashishth2018reside,peng2020learning,zhong2021frustratingly,qu2021noise}, entity types provide meaningful information for models to understand the entities, and play a crucial role in relation extraction;
    (2) \textbf{constraint information.}
    For example, it is obvious that the types of head and tail entities for relation \textit{/people/person/children} are both \textit{Person}.
    Such a constraint provides direct and effective prior information for DSRE models to recognize the noisy instances;
    (3) \textbf{interactive information.}
    As showed in Figure \ref{figure2}, relation nodes are connected indirectly through entity type nodes.
    Similar to the relation hierarchy-based methods, we can utilize message passing between nodes to transfer rich knowledge from head relations to long-tailed relations.
    
    Based on the aforementioned motivations, we propose a novel constraint graph-based relation extraction framework (CGRE).
    Our framework consists of three main components:
    a sentence encoder used for encoding the corpus information,
    a graph encoder used for encoding the constraint graph information,
    and an attention module used for integrating different information from the two separate encoders.
    Specifically, we adopt graph convolution networks(GCNs)~\cite{kipf2017semi} as the graph encoder to extract interactive information from the constraint graph.
    Intuitively, the neighborhood integration mechanism of GCNs can efficiently improve the representation learning of long-tailed relations 
    by propagating rich information from popular nodes to rare nodes.
    In addition, different from the plain selective attention~\cite{lin2016neural}, 
    which is mainly dependent on the semantic similarity between sentences and relation labels,
    our constraint-aware attention combines both the semantic information and constraint information.
    In our approach, the attention score of each instance depends on not only its semantic similarity with the relation representations,
    but also its compatibility with the entity type constraints.
    We summarize our main contributions as follows:

    \begin{itemize}[leftmargin=*]
        \item We explore a novel knowledge structure, constraint graph, 
        to model the intrinsic dependencies between relation labels.
        Compared with the popular relation hierarchical trees, 
        our constraint graph shows better universality, directness and effectiveness.
        We propose CGRE, a constraint graph-based DSRE framework, to simultaneously address the challenges of label noise and long-tailed distributions.
        \item We conduct large-scale experiments for performance comparison and ablation studies to demonstrate that CGRE is highly effective for both denoising and long-tailed RE. 
        \item We make our pre-processed datasets and source code publicly available at \href{https://github.com/tmliang/CGRE}{https://github.com/tmliang/CGRE}.
    \end{itemize}

    \begin{figure*}[ht]
      \centering
      \includegraphics[scale=0.5]{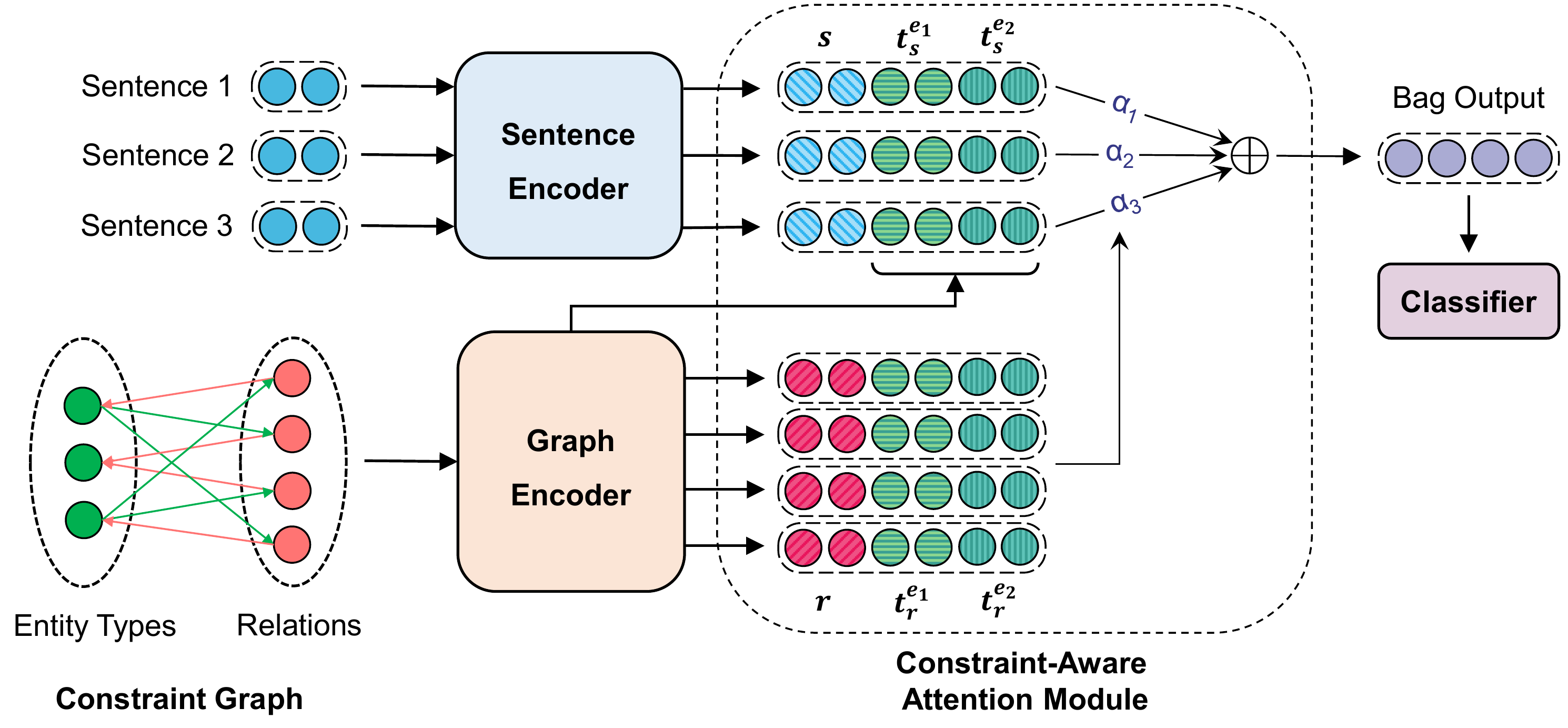}
      \caption{Overview of the proposed constraint graph-based framework.
      The sentence encoder aims to extract the representations for the sentences in a bag, 
      while the graph encoder aims to extract the representations for the relations and the entity types from the constraint graph.
      In the constraint-aware attention module, selective attention is applied after concatenation operations to aggregate the three representations into a bag representation,
      based on which the classifier predicts the relations mentioned in the sentence bag.
      }
      \label{figure3}
  \end{figure*}

    \section{Framework}
    Starting with notations and definitions, we will introduce the construction process for a raw constraint graph,
    and then detail each component of our framework.
    As illustrated in Figure \ref{figure3}, our framework consists of three key modules: 

    \begin{itemize}[leftmargin=*]
      \item \textbf{Sentence Encoder}. Given a sentence with two mentioned entities, the sentence encoder is adopted to derive a sentence representation.
    
      \item \textbf{Graph Encoder}. Given a raw constraint graph $\mathcal{G}$, we first transform it into an embedding matrix and then apply the graph encoder 
    to extract the representations of relations and entity types.

    \item \textbf{Constraint-Aware Attention}. By combining outputs from the two seperate encoders,
    this module allocates an attention score for each instance in a bag, and then computes the bag representation.
    \end{itemize}

    \subsection{Notations and Definitions}
    We define a constraint graph as a triple $\mathcal{G}=\{\mathcal{T}, \mathcal{R}, \mathcal{C}\}$, 
    where $\mathcal{T}$, $\mathcal{R}$, $\mathcal{C}$ indicate the sets of entity types, relations, and constraints, respectively.
    Each constraint $(t_{r}^{e_{1}}, r, t_{r}^{e_{2}}) \in \mathcal{C}$ indicates that for relation $r$,
    the type of head entity can be $t_{r}^{e_{1}} \in \mathcal{T}$ and the type of tail entity can be $t_{r}^{e_{2}} \in \mathcal{T}$.
    Given a bag of sentences $\mathcal{B}=\{s_{1},\ldots,s_{n_{s}}\}$ and a corresponding entity pair $(e_{1}, e_{2})$,
    the objective of distantly supervised relation extraction is to predict the relation $r_{i}$ for the entity pair $(e_{1}, e_{2})$.

    \subsection{Constraint Graph Construction}\label{sec:2.2}
    Most knowledge bases provide the entity type constraint information.
    For example, in Freebase, these constraints are located in \textit{rdf-schema\#domain} and \textit{rdf-schema\#range} fields.
    However, the original constraints always involve thousands of entity types.
    To avoid over-parameterization, we merely use 18 coarse entity types\footnotemark[1] defined in \textit{OntoNotes~5.0}~\cite{pradhan2013towards}, and then remove the constraints containing unrelated types.
    \footnotetext[1]{       
      We choose the 18 entity types of \textit{OntoNotes} for two main reasons: 
      (1) \textit{CoNLL-2003} and \textit{OntoNotes} are the two most popular named-entity recognition (NER) benchmarks. Hence, there exists a large number of \textit{CoNLL-2003}/\textit{OntoNotes}-based NER tools, which can be directly used in the instance representation construction (Sec. \ref{sec2.5.1});
      (2) the 18 types of \textit{OntoNotes} are more fine-grained than the 4 of \textit{CoNLL-2003}.
    }
    Finally, we can obtain a raw constraint graph $\mathcal{G}$ that consists of a relation set $\mathcal{R}$,
    a type set $\mathcal{T}$, and a constraint set $\mathcal{C}$. 
    Note that we use a special type \textit{Others} for the entities that do not belong to the 18 types, thus the size of $\mathcal{T}$ is 19.
    Specially, the negative category \textit{NA} (i.e., no relations between the two entities) is connected to all types, since its head/tail entities could belong to any type.
    We emphasize that the constraint graph is specific to an entire dataset, rather than a bag in the dataset. That means one public constraint graph is shared by all the bags in the dataset. 
    Furthermore, compared with the commonly used knowledge graph, the scale of a constraint graph is rather tiny. 
    For example, there are merely 72 nodes (which is the sum of relation number and type number) and 164 edges in the constraint graph for the FB-NYT dataset.

    \subsection{Sentence Encoder}
    To encode the sentence information, 
    we first adopt \textit{entity-aware word embeddings}~\cite{li2020self} to represent each word in a sentence,
    and then apply the Piecewise Convolutional Neural Network (PCNN)~\cite{zeng2015distant} to derive the sentence representation.

    \subsubsection{Input Layer}\label{2.3.1}
    The input layer aims to maps words into a distributed embedding space to capture their semantic and syntactic information.
    Given a sentence $s=\{w_{1}, \ldots, w_{l}\}$, we transform each word $w_{i}$ into a $d_{w}$-dimensional vector $\mathbf{w}_{i}$ by a pre-trained embedding matrix.
    Then following~\cite{li2020self}, we represent the target entities $e_{1}$ and $e_{2}$ by their word vectors $\mathbf{w}^{e_{1}}$ and $\mathbf{w}^{e_{2}}$.
    To incorporate the position information, 
    we use two $d_{p}$-dimensional vectors $\mathbf{p}^{e_{1}}_{i}$ and $\mathbf{p}^{e_{2}}_{i}$ to embed the relative distances between $w_{i}$ and the target entities, as used in~\cite{zeng2015distant}.
    By concatenating, two types of word embeddings can be obtained as follows:
    \begin{equation}
        \begin{aligned}
            \mathbf{x}^{p}_{i}&=[\mathbf{w}_{i} ; \mathbf{p}_{i}^{e_{1}} ; \mathbf{p}_{i}^{e_{2}}]\in\mathbb{R}^{d_{w}+2d_{p}}, \label{eq1} \\
            \mathbf{x}^{e}_{i}&=[\mathbf{w}_{i} ; \mathbf{w}^{e_{1}} ; \mathbf{w}^{e_{2}}]\in\mathbb{R}^{3d_{w}},
        \end{aligned}
    \end{equation}
    where $d_{w}$ and $d_{p}$ are both pre-defined hyper-parameters.
    Finally, we apply the entity-aware word embedding to represent each word $w_{i}$ as follows:
    \begin{equation}
        \begin{aligned}
            &\mathbf{A}^{e}=\operatorname{sigmoid}\left(\lambda \cdot\left(\mathbf{W}_{e} \mathbf{X}^{e}+\mathbf{b}_{e}\right)\right),\label{eq2} \\
            &\tilde{\mathbf{X}}^{p}=\operatorname{tanh}\left(\mathbf{W}_{p} \mathbf{X}^{p}+\mathbf{b}_{p}\right), \\
            &\mathbf{X}=\mathbf{A}^{e} \odot \mathbf{X}^{e} + \left(1-\mathbf{A}^{e}\right) \odot \tilde{\mathbf{X}}^{p},
        \end{aligned}
    \end{equation}
    where $\mathbf{X}^{p}=\{\mathbf{x}^{p}_{1}, \ldots, \mathbf{x}^{p}_{l}\}$, $\mathbf{X}^{e}=\{\mathbf{x}^{e}_{1}, \ldots, \mathbf{x}^{e}_{l}\}$,
     $\odot$ denotes element-wise product, $\mathbf{W}_{e}$ and $\mathbf{W}_{p}$ are weight matrixes, $\mathbf{b}_{e}$ and $\mathbf{b}_{p}$ are bias vectors,
    and $\lambda$ is a smoothing coefficient hyper-parameter.

    \subsubsection{Encoding Layer}
    The encoding layer aims to extract a high-dimensional representation from the input sequence.
    In consideration of simplicity and effectiveness, we employ
    PCNN~\cite{zeng2015distant} as our feature extractor.
    Given an input sequence $\mathbf{X}=\{\mathbf{x}_{1}, \ldots, \mathbf{x}_{l}\}$, 
    PCNN slides the convolutional kernels $\mathbf{W}_{k}$ over $\mathbf{X}$ to capture the hidden representations as follows:
    \begin{equation}
        \mathbf{h}_{i}=\mathbf{W}_{k}\mathbf{x}_{i-w+1:i} \in \mathbb{R}^{l} \quad 1 \leq i \leq m, 
    \end{equation}
    where $\mathbf{x}_{i:j}$ denotes the concatenating of $\mathbf{x}_{i}$ to $\mathbf{x}_{j}$, and $m$ is the number of kernels.
    Then, PCNN performs piecewise max-pooling over the hidden representations as follows:
    
    \begin{equation}
        \begin{aligned}
            \mathbf{q}^{(1)}_{i}&=\max _{1 \leq j \leq l_{1}}\left(\mathbf{h}_{ij}\right) \\
            \mathbf{q}^{(2)}_{i}&=\max _{l_{1}+1 \leq j \leq l_{2}}\left(\mathbf{h}_{ij}\right) \quad 1 \leq i \leq m ,\\
            \mathbf{q}^{(3)}_{i}&=\max _{l_{2}+1 \leq j \leq l}\left(\mathbf{h}_{ij}\right)
        \end{aligned}
    \end{equation}
    where $l_1$ and $l_2$ are positions of the two target entities respectively. 
    Then we can obtain the pooling result $\mathbf{q}_{i}=\{\mathbf{q}^{(1)}_{i}; \mathbf{q}^{(2)}_{i}, \mathbf{q}^{(3)}_{i}\}$ of the $i$-th convolutional kernel.
    Finally, we concatenate all the pooling results $\mathbf{q}_{1:m}$ and then apply a non-linear function to produce the sentence representation $\mathbf{s}$ as follows:
    \begin{equation}
        \mathbf{s}=\rho\left(\mathbf{q}_{1:m}\right)    \in\mathbb{R}^{3m},
    \end{equation}
    where $\rho(\cdot)$ is an activation function (e.g. RELU).

    Through the above-mentioned sentence encoder, 
    we can achieve the vector representation for each sentence in a bag.
    
    \subsection{Graph Encoder}
    To encode the information of the constraint graph,
    we first transform the raw graph into vector representations via the input layer,
    and then run GCNs over the input vectors to extract the interactive features of the nodes.
    Finally, the representations of entity types and relations can be obtained by dividing the node representations.

    \subsubsection{Input Layer}
    Given a raw constraint graph $\mathcal{G}=\{\mathcal{T}, \mathcal{R}, \mathcal{C}\}$,
    we denote the node set as $\mathcal{V}=\mathcal{T} \cup \mathcal{R}$.
    For each constraint $(t_{r}^{e_{1}}, r, t_{r}^{e_{2}}) \in \mathcal{C}$, we add $(t_{r}^{e_{1}}, r)$ and $(r, t_{r}^{e_{2}})$ into the edge set $\mathcal{E}$.
    Then with the edge set, the adjacency matrix $\hat{\mathbf{A}} \in \mathbb{R}^{n \times n}$ ($n=|\mathcal{V}|$) is defined as:
    \begin{equation}
        \hat{\mathbf{A}}_{ij}=\begin{cases}
            1 \quad \text{ if } (v_i, v_j) \in \mathcal{E}, \\
            0 \quad \text{otherwise}.
        \end{cases}       
    \end{equation}
    For each node $v_{i} \in \mathcal{V}$, we randomly initialize a $d_{v}$-dimensional embedding $\mathbf{v}_{i}^{(0)}$.
    Through the aforesaid process, the raw constraint graph is transformed into  
    an embedding matrix $\mathbf{V}^{(0)}=\{\mathbf{v}_{1}^{(0)}, \ldots, \mathbf{v}_{n}^{(0)}\}$ 
    with an adjacency matrix $\hat{\mathbf{A}}$. 
    
    \subsubsection{Encoding Layer}
    In this study, we select a two-layer GCN to encode the graph information.
    GCNs are neural networks that operate directly on graph structures~\cite{kipf2017semi}.
    With the neighborhood integration mechanism, GCNs can effectively promote information propagation in the graph.
    To aggregate the information of the node itself~\cite{kipf2017semi,marcheggiani2017encoding}, we add the self-loops into $\mathcal{E}$,
    which means $\hat{\mathbf{A}}_{ii}=1$.
    With the node embeddings $\mathbf{V}^{(0)}$ and the adjacency matrix $\hat{\mathbf{A}}$ as inputs, 
    we apply GCNs to extract high-dimensional representations of nodes.
    The computation for node $v_{i}$ at the $k$-th layer in GCNs can be defined as:
    \begin{equation}
        \mathbf{v}_{i}^{(k)}=\rho\left(\sum_{j=1}^{n} \hat{\mathbf{A}}_{i j} \mathbf{W}^{(k)} \mathbf{v}_{j}^{(k-1)}+\mathbf{b}^{(k)}\right),
    \end{equation}
    where $\mathbf{W}^{(k)}$ and $\mathbf{b}^{(k)}$ are respectively the weight matrix and the bias vector of the $k$-th layer, 
    and $\rho(\cdot)$ is an activation function (e.g. RELU).

    Finally, according to the category of each node,
    we divide the output representations $\mathbf{V}^{(2)} \in \mathbb{R}^{n \times d_{n}}$ 
    into relation representations $\mathbf{R} \in \mathbb{R}^{n_{r} \times d_{n}}$ and type representations $\mathbf{T} \in \mathbb{R}^{n_{t} \times d_{n}}$.

    \subsection{Constraint-Aware Attention Module}
    Given a bag of sentences $\mathcal{B}=\{s_{1},\ldots,s_{n_{s}}\}$ and a raw constraint graph $\mathcal{G}$,
    we achieved the sentence representations $\mathbf{S}=\{\mathbf{s}_{1},\ldots,\mathbf{s}_{n_{s}}\}$,
    the relation representations $\mathbf{R}=\{\mathbf{r}_{1},\ldots,\mathbf{r}_{n_{r}}\}$,
    and the type representations $\mathbf{T}=\{\mathbf{t}_{1},\ldots,\mathbf{t}_{n_{t}}\}$ by the two aforementioned encoders,
    where $n_{s}$, $n_{r}$, and $n_{t}$ are the numbers of sentences, relations, and types, respectively.
    To aggregate the information of the instances and the constraints,
    we first construct the instance representations and the constraint representations by 
    concatenation operations, and then apply the selective attention over these to compute the final bag output.

    \subsubsection{Instance Representation}\label{sec2.5.1}
    For each sentence $s_{i}$, we use NER tools\footnotemark[2] to recognize the types of its target entities.
    \footnotetext[2]{To balance the efficiency and the accuracy, we select Flair~\cite{akbik2019flair} as our NER tool.}
    For the unrecognized entities, we assign the special type \textit{Others}.
    After NER, the type representations $\mathbf{t}^{e_{1}}$ and $\mathbf{t}^{e_{2}}$ can be obtained by looking up in the type representation matrix $\mathbf{T}$.
    Finally, the instance representation is achieved by concatenating the the representations of the sentence and its entity type representations as follows:
    \begin{equation}
    \mathbf{g}_{i} = [\mathbf{s}_{i} ; \mathbf{t}_{s_{i}}^{e_{1}} ; \mathbf{t}_{s_{i}}^{e_{2}}] \in\mathbb{R}^{3d_{h}} \quad 1 \leq i \leq n_{s}.
    \end{equation}

    For example, given a sentence "\textit{Bill Gates} was 19 when he and Paul Allen started \textit{Microsoft}",
    we firstly use NER tools to recognize the types of the head entity \textit{Bill Gates} and the tail entity \textit{Microsoft} as \textit{Person} and \textit{Organization} respectively.
    Then the instance representation is achieved as 
    $\mathbf{g} = [\mathbf{s} ; \mathbf{t}_{per} ; \mathbf{t}_{org}]$,
    where $\mathbf{s}$ is the sentence representation, 
    $\mathbf{t}_{per}$ is the representation of \textit{Person},
    and $\mathbf{t}_{org}$ is the representation of \textit{Organization}.

    \subsubsection{Constraint Representation}\label{sec2.5.2}
    Assume $t^{e_{1}}_{r_{i}}$ and $t^{e_{2}}_{r_{i}}$ are respectively the immediate predecessor and the immediate successor of relation $r_{i}$ in the constraint graph.
    Similarly, we can obtain the type representations $\mathbf{t}^{e_{1}}_{r_{i}}$ and $\mathbf{t}^{e_{2}}_{r_{i}}$ of $r_{i}$ by looking up in $\mathbf{T}$.
    Note that some relations may have multiple immediate predecessors or successors,
    e.g. the tail entity type of \textit{/location/location/contains} can be \textit{GPE} (which denotes countries, cities, and states) 
    or \textit{LOC} (which denotes non-GPE locations such as mountains and rivers).
    For these cases, we take the average over the immediate predecessors or the immediate successors.
    Finally, the constraint representation is achieved by concatenating the representations of the relation and its entity type representations as
    \begin{equation}
    \mathbf{c}_{i} = [\mathbf{r}_{i} ; \mathbf{t}_{r_{i}}^{e_{1}} ; \mathbf{t}_{r_{i}}^{e_{2}}] \in\mathbb{R}^{3d_{h}} \quad 1 \leq i \leq n_{r}.
    \end{equation}

    For example, assume that \textit{Person} and \textit{Organization} are 
    respectively the unique immediate predecessor and the unique immediate successor 
    of \textit{/business/person/company} in the constraint graph,
    then the constraint representation of \textit{/business/person/company} is achieved as 
    $\mathbf{c} = [\mathbf{r} ; \mathbf{t}_{per} ; \mathbf{t}_{org}]$,
    where $\mathbf{r}$ is the representation of \textit{/business/person/company}, 
    $\mathbf{t}_{per}$ is the representation of \textit{Person},
    and $\mathbf{t}_{org}$ is the representation of \textit{Organization}.

    \subsubsection{Attention Layer}
    Different from the previous selective attention mechanisms that mainly depend on the semantic similarity between sentences and relations~\cite{lin2016neural,alt2019fine},
    our attention mechanism combines semantic information and constraint information to calculate the bag output.
    Intuitively, the similarity between the former parts (i.e., $\mathbf{s}$ and $\mathbf{r}$) measures the semantic matching between the sentence $s$ and the relation $r$, while that between the latter parts (i.e., $[\mathbf{t}_{s}^{e_{1}} ; \mathbf{t}_{s}^{e_{2}}]$ and $[\mathbf{t}_{r}^{e_{1}} ; \mathbf{t}_{r}^{e_{2}}]$) measures agreement of the instance with the constraint of $r$. 
    Given a bag of instances, the attention weight $\alpha_{i}$ of $i$-th instance for the corresponding relation $r$ can be computed as follows:
    \begin{equation}
        \begin{aligned}
            e_{i}&=\mathbf{g}_{i} \mathbf{c}_{r}, \\
            \alpha_{i}&=\frac{\exp \left(e_{i}\right)}{\sum_{j=1}^{n_{s}} \exp \left(e_{j}\right)},
        \end{aligned}
    \end{equation}
    where $\mathbf{c}_{r}$ is the constraint representation of $r$. 
    Then the bag representation is derived as the weighted sum of the sentence representations:
    \begin{equation}
        \mathbf{z}_{r}=\sum_{i=1}^{n_{s}} \alpha_{i} \mathbf{g}_{i}.
    \end{equation}
   
    Finally, we feed the bag representation $\mathbf{z}_{r}$ into a softmax classifier to calculate the probability distribution over relation labels as follows:
    \begin{equation}
        P\left(r \mid \mathcal{B};\mathcal{G};\theta\right)=\operatorname{softmax}\left(\mathbf{W}\mathbf{z}_{r}+\mathbf{b}\right), 
    \end{equation}
    where $\theta$ is the set of model parameters, $\mathbf{W}$ is the weight of the classifier and $\mathbf{b}$ is the bias.

    \subsection{Optimization}
    We define the objective function using cross-entropy at the bag level as follows:
    \begin{equation}
        J(\theta)=-\frac{1}{n} \sum_{i=1}^{n} \log P\left(r_{i} \mid \mathcal{B}_{i} ; \mathcal{G} ; \theta\right),   
    \end{equation}
    where $n$ is the number of bags and $r_{i}$ is the label of $\mathcal{B}_{i}$.
    
    Note that at the test stage, the ground-truth label $r$ is unknown, 
    thus all constraint representations are applied to calculate the posterior probabilities for the corresponding relation,
    and the relation with the highest probability is the prediction result~\cite{lin2016neural}.

    \section{Experiments}
    \subsection{Datasets}
    \begin{table}[t]
      \centering
      \caption{Statistics for NYT-520K, NYT-570K and GDS datasets, where
      \textit{Ins.} and \textit{Ent.} stand for instances and entity pairs respectively.}
      \begin{tabular}{cccccc}
          \toprule
          \multirow{2}*{\textbf{Dataset}} & \multicolumn{2}{c}{\textbf{Training}} & \multicolumn{2}{c}{\textbf{Test}} & \multirow{2}*{\textbf{\# Relation}} \\
          \cmidrule(lr){2-3} \cmidrule(lr){4-5}
            ~ & \# \textit{Ins.} & \# \textit{Ent.} & \# \textit{Ins.} & \# \textit{Ent.} & ~ \\
          \midrule
          NYT-520K & 523,312 & 280,275 & \multirow{2}*{172,448} & \multirow{2}*{96,678} & \multirow{2}*{53} \\
          NYT-570K & 570,088 & 291,699 & ~ & ~ & ~ \\
          \midrule
          GDS & 13,161 & 7,580 & 5,663 & 3,247 & 5 \\
          \bottomrule
        \end{tabular}
        \label{table1}
    \end{table}

    We use two popular benchmark datasets for our primary experiments:

    \begin{itemize}[leftmargin=*]
      \item \textbf{FB-NYT}~\cite{riedel2010modeling} is generated by aligning Freebase facts with a New York Times corpus.
      As described in \cite{bai2019structured} and \cite{christopoulou2021distantly}, FB-NYT has been released in two main versions: 
      the filtered version \textbf{NYT-520K} and the non-filtered version \textbf{NYT-570K}. The two versions are identical, except the training set of NYT-520K does not share any entity pairs with the test set. NYT-570K has been widely used in previous studies\cite{lin2016neural,vashishth2018reside,han2018hierarchical,ye2019distant}, while NYT-520K provides a more accurate evaluation for model's in-depth comprehension of the relation semantics rather than superficial memory of the KB facts during training \cite{shang2020noisy,zhang2020robust,cao2021learning,qu2021noise}. Therefore, we experiment with both the two versions of FB-NYT.

      \item \textbf{GDS}\cite{jat2018improving} is constructed from human-annotated Google Relation Extraction corpus with additional instances from Internet documents, guaranteeing that each bag contains at least one sentence that expresses the bag label. The guarantee enhances the reliability for bag-level automatic evaluation, hence GDS has become a popular benchmark in recent DSRE studies based on multi-instance learning\cite{vashishth2018reside,cao2021learning,qu2021noise,hao2021knowing}.
    \end{itemize}
    We present the overall statistics for NYT-520K, NYT-570K and GDS in Table \ref{table1}. 

    \subsection{Hyper-parameters Settings}
    For a fair comparison, most of the hyper-parameters are set identical to those in ~\cite{lin2016neural},
    and we mainly tune the hyper-parameters of the graph encoder and the classifier.
    The word embeddings are initialized by the pre-trained word2vec released by OpenNRE\footnotemark[3]. 
    \footnotetext[3]{\href{https://github.com/thunlp/OpenNRE}{https://github.com/thunlp/OpenNRE}}
    All weight matrixes and position embeddings are initialized by Xavier initialization~\cite{glorot2010understanding},
    and the bias vectors are all initialized to 0.
    To prevent over-fitting, we apply dropout~\cite{srivastava2014dropout} before the classifier layer.
    Table \ref{table2} lists all the hyper-parameters used in our experiments.

    \begin{table}[t]
      \centering
      \caption{Hyper-parameters settings for NYT-520K, NYT-570K and GDS.}
      \begin{tabular}{clccc}
          \toprule
          \textbf{Component} & \multicolumn{1}{c}{\textbf{Parameters}} & \textbf{NYT-520K} & \textbf{NYT-570K} & \textbf{GDS} \\
          \midrule
          \multirow{5}*{\shortstack{Sentence\\Encoder}} & filter num. & 230 & 230 & 230 \\
          ~ & window size & 3 & 3 & 3 \\
          ~ & word size & 50 & 50 & 50 \\
          ~ & position size & 5 & 5 & 5 \\
          ~ & coefficient $\lambda$ & 17 & 20 & 17 \\
           \midrule
          \multirow{3}*{\shortstack{Graph\\Encoder}} & emb. size & 100 & 700 & 150 \\ 
          ~ & hidden size & 750 & 950 & 900 \\
          ~ & output size & 1150 & 690 & 150 \\
          \midrule
          Classifier & input size & 650 & 690 & 300 \\
          \midrule
          \multirow{3}*{Optimization} & batch size & 160 & 160 & 160 \\
          ~ & learning rate & 0.5 & 0.5 & 0.5 \\
          ~ & dropout rate & 0.5 & 0.5 & 0.5 \\
           \bottomrule
      \end{tabular}
      \label{table2}
  \end{table}

    \subsection{Denoising Evaluation}\label{Denoising Evaluation}
    To demonstrate the denoising performance of CGRE,
    we compare against five competitive DSRE models:
    \begin{itemize}[leftmargin=*]
      \item \textbf{PCNN+ATT}~\cite{lin2016neural}: a PCNN-based model with selective attention;
      \item \textbf{PCNN+HATT}~\cite{han2018hierarchical}: a PCNN-based model with hierarchical attention;
      \item \textbf{PCNN+BATT}~\cite{ye2019distant}: a PCNN-based model with intra-bag and inter-bag attention;     
      \item \textbf{RESIDE}~\cite{vashishth2018reside}: a DSRE model integrating the external information including relation alias and entity type.
      \item \textbf{DSRE-VAE}~\cite{christopoulou2021distantly}: a DSRE model based on Variational Autoencoder (VAE), which could be further improved by incorporating external KB priors.
    \end{itemize}
    
    Note that the first four models (i.e., PCNN+ATT\footnotemark[3], PCNN+HATT\footnotemark[4], PCNN+BATT\footnotemark[5] and RESIDE\footnotemark[6]) were originally applied on NYT-570K.
    Hence, their results on NYT-570K are extracted from the respective publications.
    For DSRE-VAE\footnotemark[7], results on both NYT-520K and NYT-570K are reported in the original publication.
    Other results were obtained using the official source codes. 
    We do not test DSRE-VAE (+KB) on the GDS dataset, since the KB of GDS is unavailable.
    \footnotetext[4]{\href{https://github.com/thunlp/HNRE}{https://github.com/thunlp/HNRE}}
    \footnotetext[5]{\href{https://github.com/ZhixiuYe/Intra-Bag-and-Inter-Bag-Attentions}{https://github.com/ZhixiuYe/Intra-Bag-and-Inter-Bag-Attentions}}
    \footnotetext[6]{\href{https://github.com/malllabiisc/RESIDE}{https://github.com/malllabiisc/RESIDE}} 
    \footnotetext[7]{\href{https://github.com/fenchri/dsre-vae}{https://github.com/fenchri/dsre-vae}}

    To further evaluate the effectiveness of entity-aware word embedding proposed by \cite{li2020self}, 
    we additionally build \textbf{PCNN+ATT+ENT} as a baseline by simply replacing the input layer of PCNN+ATT with the entity-aware word embedding layer (as described in Sec. \ref{2.3.1}).

\begin{figure*}[t]
    \centering
    \subfigure[NYT-520K]{\includegraphics[scale=0.36]{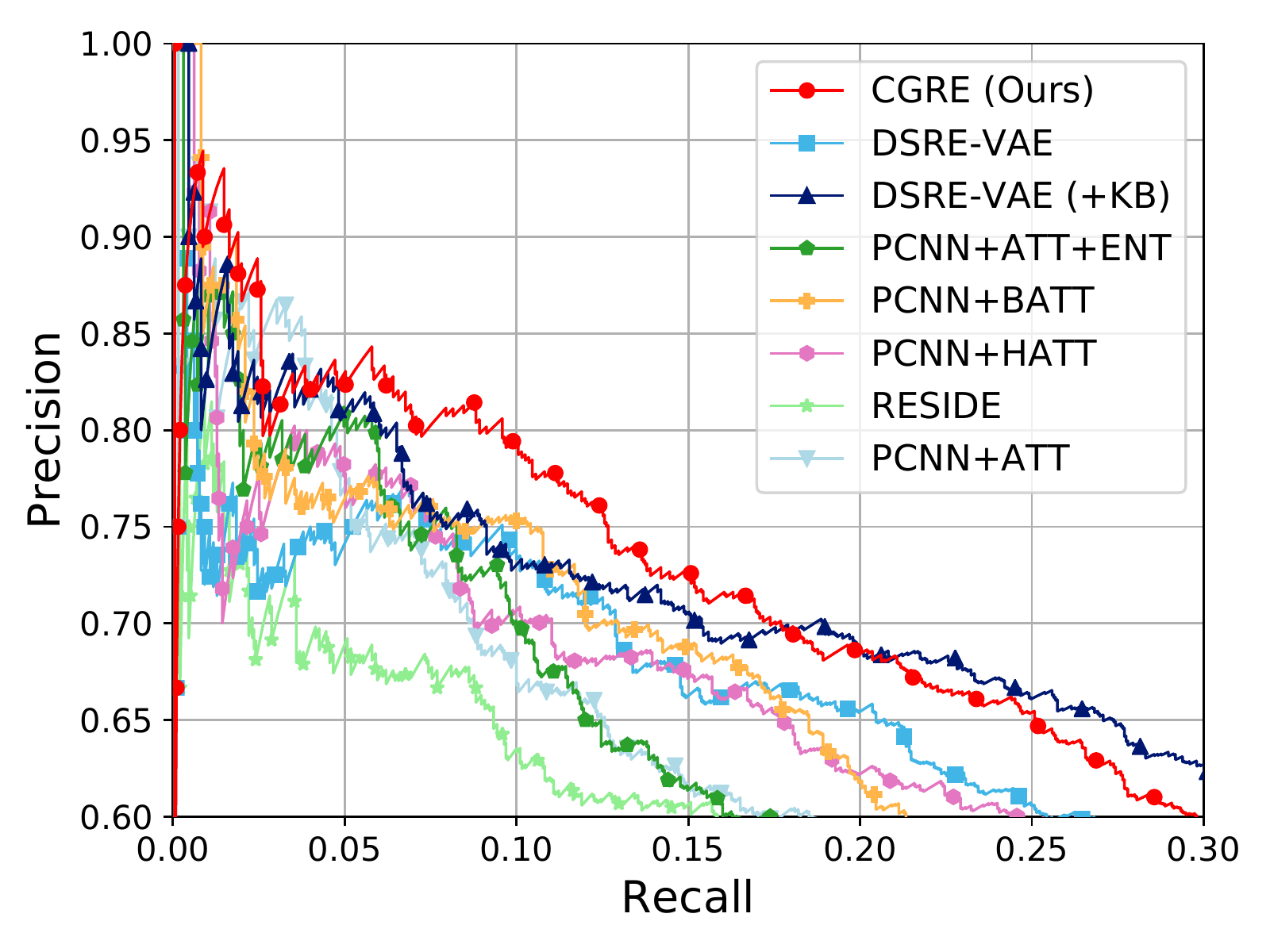}\label{figure4.1}}
    \subfigure[NYT-570K]{\includegraphics[scale=0.36]{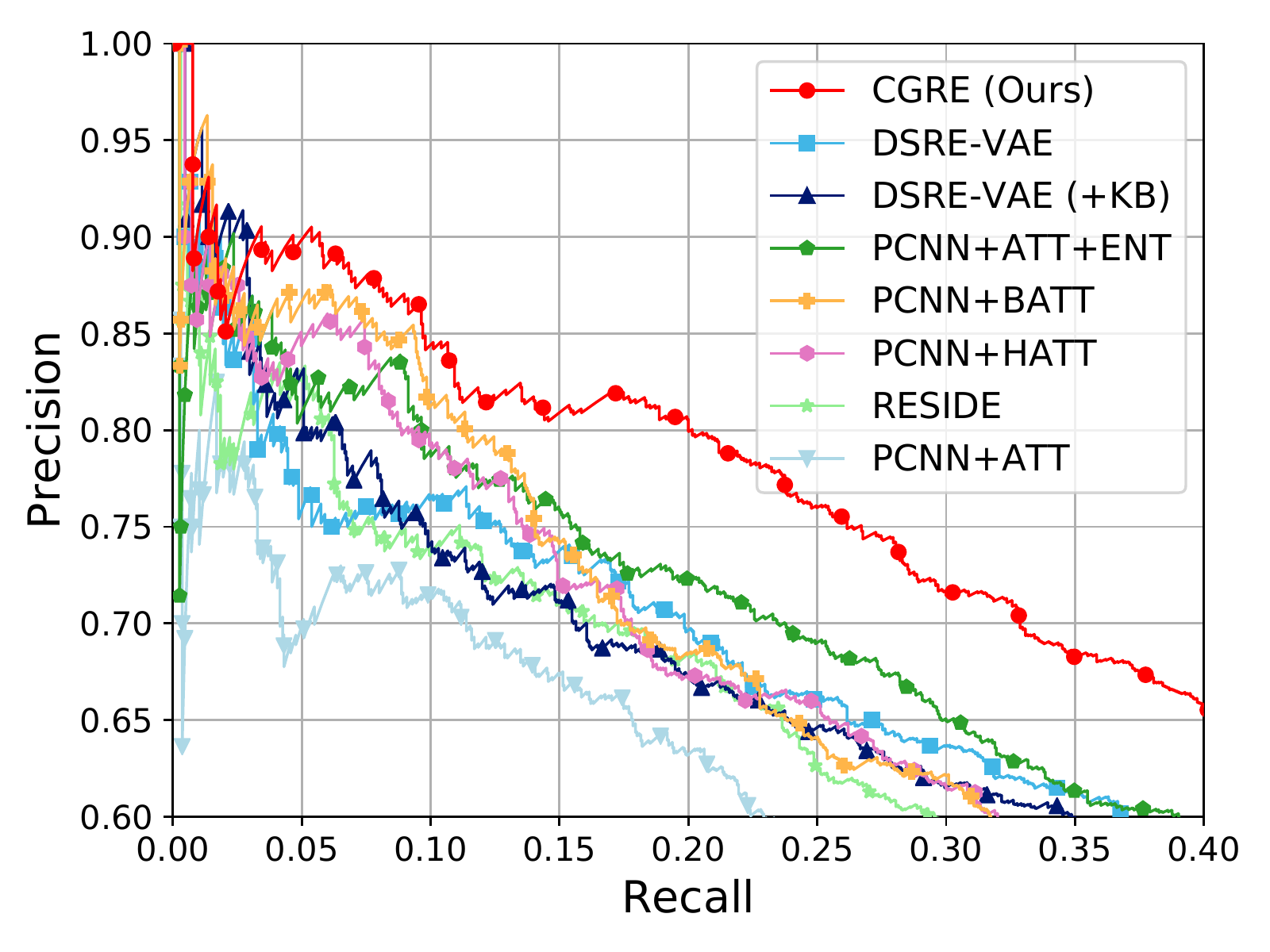}\label{figure4.2}}
    \subfigure[GDS]{\includegraphics[scale=0.36]{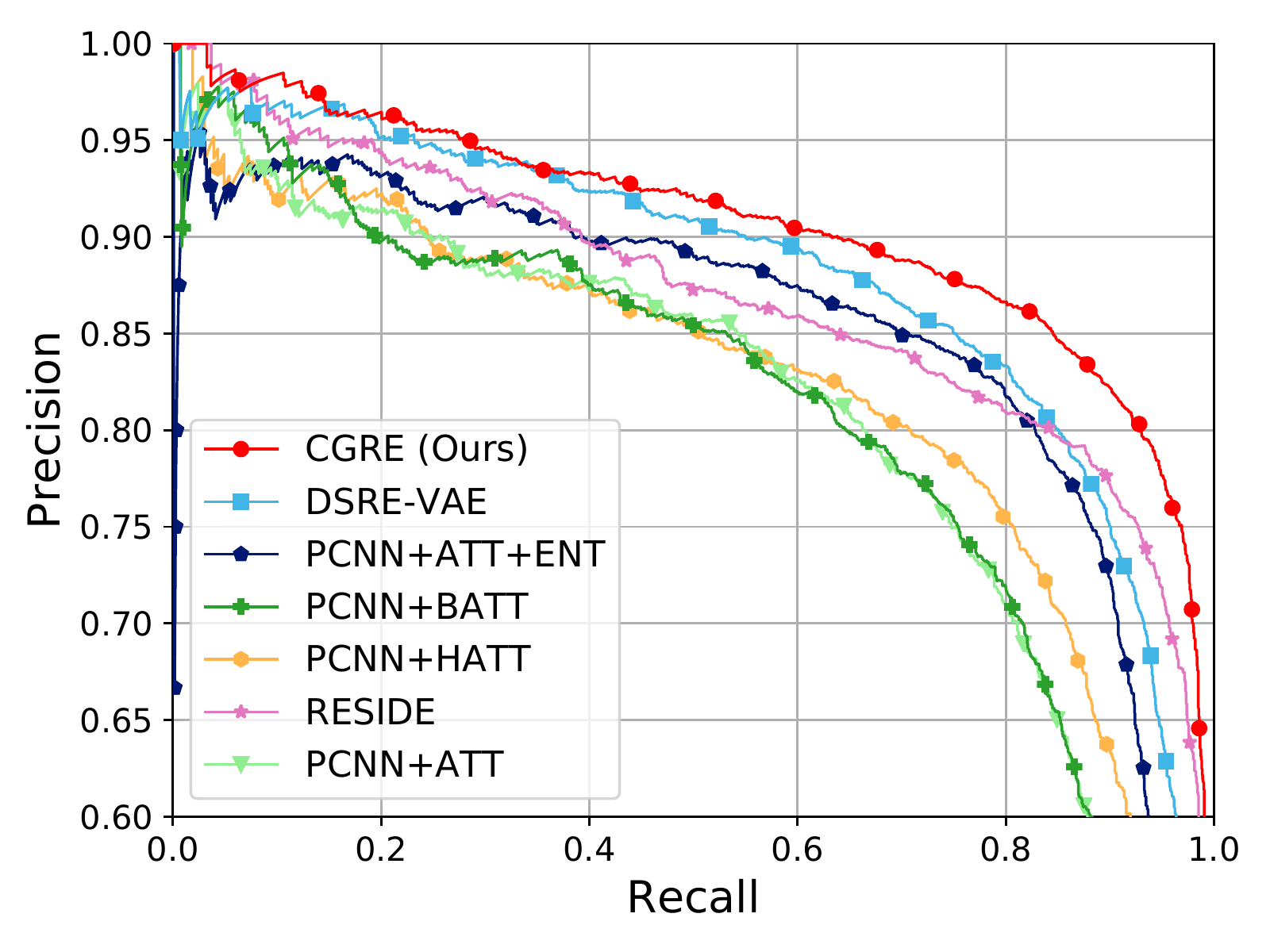}\label{figure4.3}}
    \caption{PR curves for different models on NYT-520K, NYT-570K and GDS.
    Note that \textit{DSRE-VAE} indicates the vanilla DSRE-VAE, and \textit{DSRE-VAE (+KB)} indicates the the DSRE-VAE incorporating KB priors.
    }\label{figure4}
\end{figure*}

    Following the previous works~\cite{mintz2009distant, lin2016neural,han2018hierarchical, vashishth2018reside, ye2019distant}, 
    we adopt precision-recall (PR) curves shown in Figure \ref{figure4} 
    to measure the overall performance of DSRE models in a noisy environment.
    For a ready comparison, we report the accuracy of top-N predictions (P@N) and the area under curve (AUC) of the PR curves in Table \ref{table3}. 
    
    Our observations on the denoising results shown in Figure \ref{figure4} and Table \ref{table3} are summarized as follows:

    (1) \textbf{CGRE vs. PCNN+ATTs.} 
    As a variant of PCNN+ATT, CGRE improves the performance of vanilla PCNN+ATT by a large margin. As observed in Table \ref{table3}, CGRE outperforms other PCNN+ATT variants, i.e. PCNN+HATT and PCNN+BATT, by at least 4.6\%, 9.9\% and 8.9\% on NYT-520K, NYT-570K and GDS, respectively;

    (2) \textbf{CGRE vs. RESIDE.}
    CGRE shows high efficiency and sufficiency in utilizing type information. Compared with RESIDE which uses 38 entity types\footnotemark[8]\footnotetext[8]{RESIDE integates 38 entity types defined in the first hierarchy of FIGER\cite{ling2012fine}.}, CGRE merely uses 18 types but achieves AUC improvements of 6.2\%, 10.4\% and 2.7\% on NYT-520K, NYT-570K and GDS, respectively. Furthermore, Figure \ref{figure4.3} and Table \ref{table3} also demonstrate the significance of utilizing type information in situations with scarce training data, as CGRE and RESIDE show considerable improvements on GDS (whose training data is less than 3\% of FB-NYT's);

    (3) \textbf{CGRE vs. DSRE-VAEs.}
    On NYT-520K, CGRE outperforms the vanilla DSRE-VAE by 3.1\% in AUC values, but underperforms DSRE-VAE (+KB) by 1.2\%. We believe the performance gap between CGRE and DSRE-VAE (+KB) arises from the different external information they use. Seemingly, the KB priors (knowledge graph embeddings) in DSRE-VAE (+KB) contribute higher performance improvements than the constraint graph in CGRE. However, note that the scale of the former is several orders of magnitude larger than that of the latter. For the case of NYT-520K, the knowledge graph used by DSRE-VAE (+KB) contains 3,065,045 nodes and 24,717,676 edges, while the constraint graph used by CGRE merely contains 72 nodes and 164 edges (as mentioned in Sec \ref{sec:2.2}). Nevertheless, CGRE can still achieve performance comparable with DSRE-VAE (+KB) on NYT-520K, and on NYT-570K, CGRE outperforms DSRE-VAE (+KB) by a large margin.
    These comparison results demonstrate the effectiveness and promise of our proposed Constraint Graph and CGRE. 
    In view of the performance advantage of DSRE-VAE (+KB), integrating the KB priors into CGRE for further improvements is worthy of exploration in future works.

    \begin{table*}[t]
      \centering
      \caption{(\%) P@N and AUC values of different models on NYT-520K, NYT-570K and GDS.}
      \begin{tabular}{lcccccccccccc}
          \toprule
          \multirow{2}*{\textbf{Model}} & \multicolumn{4}{c}{\textbf{NYT-520K}} & \multicolumn{4}{c}{\textbf{NYT-570K}} & \multicolumn{4}{c}{\textbf{GDS}} \\
          \cmidrule(lr){2-5}\cmidrule(lr){6-9}\cmidrule(lr){10-13}
          ~ & P@100 & P@200 & P@300 & AUC & P@100 & P@200 & P@300 & AUC & P@100 & P@200 & P@300 & AUC \\
          \midrule
          PCNN+ATT~\cite{lin2016neural} & 81.6 & 73.0 & 66.9 & 34.7 & 72.9 & 71.5 & 69.6 & 38.4 & 96.4 & 93.3 & 91.5 & 79.9 \\
          PCNN+HATT~\cite{han2018hierarchical} & 77.5 & 74.2 & 69.9 & 37.1 & 85.0 & 81.5 & 77.5 & 41.9 & 94.0 & 92.9 & 92.4 & 81.6\\
          PCNN+BATT~\cite{ye2019distant} & 76.9 & 75.4 & 72.9 & 35.2 & 86.5 & 84.7 & 79.4 & 42.0 & 96.3 & 94.7 & 93.1 & 80.2 \\
          RESIDE~\cite{vashishth2018reside} & 69.3 & 67.5 & 63.4 & 35.5 & 81.4 & 74.9 & 73.8 & 41.5 & 98.4 & 96.4 & 95.2 & 86.8\\
          DSRE-VAE~\cite{christopoulou2021distantly} & 74.0 & 74.5 & 71.6 & 38.6 & 80.0 & 76.0 & 75.6 & 44.6 & 96.9 & 96.7 & 96.3 & 87.6 \\
          DSRE-VAE (+KB)~\cite{christopoulou2021distantly} & \textbf{83.0} & 75.5 & 73.0 & \textbf{42.9} & 81.0 & 77.5 & 73.6 & 45.5 & - & - & - & - \\
          \midrule
          PCNN+ATT+ENT & 79.4 & 74.7 & 68.3 & 35.5 & 83.7 & 81.4 & 77.5 & 47.0 & 93.9 & 93.7 & 93.5 & 84.9 \\
          \midrule
          CGRE (Ours) & 82.7 & \textbf{80.3} & \textbf{76.5} & 41.7 & \textbf{88.9} & \textbf{86.4} & \textbf{81.8} & \textbf{51.9} & \textbf{98.6} & \textbf{98.1} & \textbf{97.7} & \textbf{90.5} \\
          \bottomrule
      \end{tabular}
      \label{table3}
    \end{table*}

    \subsection{Long Tail Evaluation}\label{sec:3.4}
    To fully demonstrate the effectiveness of CGRE on long-tail RE,
    we compare CGRE with several competitive baselines, 
    including two vanilla DSRE models \textbf{PCNN+ATT}~\cite{lin2016neural} and \textbf{PCNN+ATT+ENT} that ignore the long-tailed relations,
    and six state-of-the-art long-tailed RE models:
    \begin{itemize}[leftmargin=*]
      \item \textbf{PCNN+HATT}~\cite{han2018hierarchical}: the first model to apply relation hierarchies for long-tailed RE;
      \item \textbf{PCNN+KATT}~\cite{zhang2019long}: a model that utilizes GCNs to encode relation hierarchies and applies knowledge graph embeddings for initialization;
      \item \textbf{PA-TRP}~\cite{cao2021learning}: a model that learns relation prototypes from unlabeled texts for embedding relation hierarchies;
      \item \textbf{CoRA}~\cite{li2020improving}: a model that uses sentence embeddings to generate the representations for each node in relation hierarchies;
      \item \textbf{ToHRE}~\cite{yu2020tohre}: a model that applies a top-down classification strategy to explore relation hierarchies;
      \item \textbf{DPEN}~\cite{gou2020dynamic}: a model that aggregates relations and entity types by stages to dynamically enhance the classifier.
    \end{itemize}

    Note that the first five models are all based on the relation hierarchy.

    \begin{table}[t]
      \centering
      \caption{(\%) Accuracy of Hits@K on relations with training instances fewer than 100/200.}
      \begin{tabular}{lcccccc}
          \toprule
          \textbf{Training Instances} & \multicolumn{3}{c}{\textbf{$<$100}} & \multicolumn{3}{c}{\textbf{$<$200}} \\
          \cmidrule(lr){2-4}\cmidrule(lr){5-7}
          \textbf{Hits@K (Macro)} & 10 & 15 & 20 & 10 & 15 & 20 \\
          \midrule
              PCNN+ATT~\cite{lin2016neural} & $<$5.0 & 7.4 & 40.7 & 17.2 & 24.2 & 51.5 \\
              PCNN+ATT+ENT & 22.2 & 33.3 & 61.1 & 36.4 & 45.5 & 68.2 \\
          \midrule
              PCNN+HATT~\cite{han2018hierarchical} & 29.6 & 51.9 & 61.1 & 41.4 & 60.6 & 68.2 \\
              PCNN+KATT~\cite{zhang2019long} & 35.3 & 62.4 & 65.1 & 43.2 & 61.3 & 69.2 \\
              PA-TRP~\cite{cao2021learning} & 63.9 & 70.3 & 72.2 & 66.7 & 72.3 & 73.8 \\
              CoRA~\cite{li2020improving} & 59.7 & 63.9 & 73.6 & 65.4 & 69.0 & 77.4 \\
              ToHRE~\cite{yu2020tohre} & 62.9 & 75.9 & 81.4 & 69.7 & 80.3 & 84.8 \\
          \midrule
              DPEN~\cite{gou2020dynamic} & 57.6 & 62.1 & 66.7 & 64.1 & 68.0 & 71.8 \\
          \midrule
              CGRE (Ours) & \textbf{77.8} & \textbf{77.8} & \textbf{87.0} & \textbf{81.8} & \textbf{81.8} & \textbf{89.4} \\
          \bottomrule
      \end{tabular}
      \label{table4}
    \end{table}

    Following the previous studies~\cite{han2018hierarchical,zhang2019long}, 
    we adopt Hit@K metric on \textbf{NYT-570K} to evaluate the performance on long-tailed relations.
    We extract the relations that have few than 100/200 training instances from the test set.
    Then Hits@K metric, which measures the probability that the true label falls in the top-$K$ recommendations of the model,
    is applied over these long-tailed relations.
    From the results in Table \ref{table4}, we observe:
    
    (1) \textbf{effectiveness of modeling latent dependencies among relations.}
    PCNN+ATT and PCNN+ATT+ENT give the worst results, showing the ineffectiveness of vanilla denoising models on long-tailed relations.
    By modeling the latent dependencies among relations, the performance improves significantly, but is still far from satisfactory.
    
    (2) \textbf{advantages of the constraint graph over the relation hierarchical tree.} 
    Based on the constraint graph, CGRE significantly and consistently outperforms all the previous relation hierarchy-based models.
    This confirms the advantages of the constraint graph for modeling dependencies among relations, which could be further amplified via the GCN.

    (3) \textbf{superiority of CGRE over DPEN in utilizing type information.}
    As compared with DPEN, which also utilizes entity types to transfer information across relations, CGRE achieves significant and consistent performance superiority for long-tailed RE. 
    We believe this performance gap stems mainly from the distinction in how the methods model and utilize type information.
    DPEN models the connections between entity types and relations implicitly, while CGRE makes these explicit via the constraint graph. 
    Also, DPEN aggregates the features of entity types and relations by stages with a specific dynamic parameter generator, while CGRE adopts GCNs to achieve this objective directly.

    As described in the aforementioned observations,
    long-tailed RE is still an intractable challenge even for the state-of-art DSRE models.
    A feasible solution is modeling intrinsic dependencies between the relations to transfer information from data-rich relations to data-poor relations, and the design of relation dependency structure is one of the key to modeling.
    With outstanding performance, the relation hierarchy seems to be the indispensable foundation of long-tailed RE frameworks~\cite{han2018hierarchical,zhang2019long,cao2021learning,li2020improving,yu2020tohre}, however, \textit{it is not the only option}.
    In this work, we confirm the constraint graph is also a novel and effective structure for modeling latent dependencies between relations. 
    Further, we believe there are still many promising relation dependency structures deserving extensive exploration.

    \begin{table*}[ht]
      \centering
      \caption{(\%) P@N on NYT-520K with different bag sizes.}
      \begin{tabular}{lcccccccccccc}
          \toprule
          \textbf{Bag Size} & \multicolumn{4}{c}{\textbf{One}} & \multicolumn{4}{c}{\textbf{Two}} & \multicolumn{4}{c}{\textbf{All}}\\
          \cmidrule(lr){2-5}\cmidrule(lr){6-9}\cmidrule(lr){10-13}
          \textbf{P@N} & 100 & 200 & 300 & Mean & 100 & 200 & 300 & Mean & 100 & 200 & 300 & Mean \\
          \midrule
          PCNN+ATT~\cite{lin2016neural} & 74.0 & 65.0 & 61.3 & 66.8 & 78.0 & 69.0 & 65.7 & 70.9 & 83.0 & 69.5 & 67.0 & 73.2 \\
          PCNN+HATT~\cite{han2018hierarchical} & 75.0 & 70.0 & 66.0 & 70.3 & 76.0 & 73.5 & 67.7 & 72.4 & 77.0 & 73.5 & 69.0 & 73.2\\
          PCNN+BATT~\cite{ye2019distant} & 74.0 & 65.0 & 63.7 & 67.6 & 81.0 & 72.0 & 67.4 & 72.6 & 86.0 & 75.5 & 68.3 & 76.6\\
          RESIDE~\cite{vashishth2018reside} & 68.0 & 67.5 & 63.3 & 66.3 & 65.0 & 61.5 & 58.0 & 61.5 & 70.0 & 65.5 & 58.0 & 64.5\\
          DSRE-VAE~\cite{christopoulou2021distantly} & 79.0 & 70.0 & 63.3 & 70.6 & 77.0 & 70.0 & 66.0 & 71.0 & 79.0 & 75.5 & 69.0 & 74.5\\
          DSRE-VAE (+KB)~\cite{christopoulou2021distantly} & 85.0 & 73.5 & 67.0 & 75.2 & 88.0 & \textbf{77.0} & 69.7 & 78.2 & \textbf{92.0} & \textbf{80.5} & 73.0 & \textbf{81.8}\\
          \midrule
          PCNN+ATT+ENT & 78.0 & 68.0 & 66.0 & 70.7 & 79.0 & 70.0 & 64.0 & 71.0 & 79.0 & 70.5 & 66.7 & 72.1\\
          \midrule
          CGRE (Ours) & \textbf{86.0} & \textbf{74.5} & \textbf{68.0} & \textbf{76.2} & \textbf{88.0} & 76.0 & \textbf{71.7} & \textbf{78.6} & 88.0 & 77.5 & \textbf{73.3} & 79.6\\
          \bottomrule
      \end{tabular}
      \label{table5}
  \end{table*}

  \begin{table*}[ht]
    \centering
    \caption{(\%) P@N on NYT-570K with different bag sizes.}
    \begin{tabular}{lcccccccccccc}
        \toprule
        \textbf{Bag Size} & \multicolumn{4}{c}{\textbf{One}} & \multicolumn{4}{c}{\textbf{Two}} & \multicolumn{4}{c}{\textbf{All}}\\
        \cmidrule(lr){2-5}\cmidrule(lr){6-9}\cmidrule(lr){10-13}
        \textbf{P@N} & 100 & 200 & 300 & Mean & 100 & 200 & 300 & Mean & 100 & 200 & 300 & Mean \\
        \midrule
        PCNN+ATT~\cite{lin2016neural} & 73.3 & 69.2 & 60.8 & 67.8 & 77.2 & 71.6 & 66.1 & 71.6 & 76.2 & 73.1 & 67.4 & 72.2 \\
        PCNN+HATT~\cite{han2018hierarchical} & 84.0 & 76.0 & 69.7 & 76.6 & 85.0 & 76.0 & 72.7 & 77.9 & 88.0 & 79.5 & 75.3 & 80.9\\
        PCNN+BATT~\cite{ye2019distant} & 86.8 & 77.6 & 73.9 & 79.4 & 91.2 & 79.2 & 75.4 & 81.9 & 91.8 & 84.0 & 78.7 & 84.8\\
        RESIDE~\cite{vashishth2018reside} & 80.0 & 75.5 & 69.3 & 74.9 & 83.0 & 73.5 & 70.6 & 75.7 & 84.0 & 78.5 & 75.6 & 79.4\\
        DSRE-VAE~\cite{christopoulou2021distantly} & 79.0 & 68.5 & 65.7 & 71.1 & 74.0 & 72.5 & 69.3 & 71.9 & 77.0 & 72.5 & 67.3 & 72.3\\
        DSRE-VAE (+KB)~\cite{christopoulou2021distantly} & 79.0 & 72.0 & 65.7 & 72.2 & 81.0 & 74.0 & 66.7 & 73.9 & 88.0 & 74.0 & 70.7 & 77.6\\
        \midrule
        PCNN+ATT+ENT & 87.0 & 84.0 & 80.7 & 83.9 & 89.0 & 85.5 & 81.3 & 85.3 & 91.0 & 87.0 & 83.3 & 87.1\\
        \midrule
        CGRE (Ours) & \textbf{95.0} & \textbf{88.5} & \textbf{85.0} & \textbf{89.5} & \textbf{95.0} & \textbf{90.0} & \textbf{84.7} & \textbf{89.9} & \textbf{94.0} & \textbf{92.5} & \textbf{88.3} & \textbf{91.6}\\
        \bottomrule
    \end{tabular}
    \label{table6}
\end{table*}

    \subsection{Quantitative Analysis of Selective Attention Mechanisms}
    In this section, we quantitatively evaluate the effects of selective attention mechanism on DSRE, to demonstrate the performance of our designed \textit{constraint-aware attention mechanism}.

    \subsubsection{Performance of Selective Attention Mechanisms}
    We conduct top-N evaluations on the bags containing different number of sentences, to compare the performance of different selective attention mechanisms.
    Following the previous works~\cite{lin2016neural, vashishth2018reside, han2018hierarchical, ye2019distant}, we randomly select \textit{one} / \textit{two} / \textit{all} sentences for each test entity pair with more than two sentences to construct three new test sets, on which we calculate P@N for these methods.

    We report the results on NYT-520K and NYT-570K in Table \ref{table4} and \ref{table5}.
    For the bags of size \textit{one}, these selective attention mechanisms are ineffective, since they operate at the sentence level. 
    As the bag size increases, selective attention mechanisms provide increasing performance improvements of bag-level DSRE.
    Across different bag sizes, CGRE achieves performance comparable with DSRE-VAE (+KB), and significantly outperforms other selective attention-based models. 
    These results demonstrate the effectiveness of our constraint-aware attention mechanism.

    \subsubsection{Evaluation with Different Noise Ratios}
    We further quantitatively evaluate the performance of the proposed constraint-aware attention mechanism on bags with different noise ratios, using the analytical framework proposed by \cite{hu2021knowledge}. 
    
    This analytical framework requires label-clean test data for quantitative evaluation. In this work, instead of creating a new dataset from scratch, we experiment with the existing \textbf{NYT-H} dataset because it provides human-annotated test data. NYT-H consists of a training set containing 96,831 instances (14,012 entity pairs) and a test set containing 9,955 instances (3,548 entity pairs), involving 21 relations (excluding NA).
    
    In this experiment, we adopt two evaluation metrics as follows:

    \begin{itemize}[leftmargin=*]
      \item \textbf{Attention Accuracy (AAcc)}. The key goal of the selective attention mechanism is to dynamically assign higher weights to valid sentences and lower weights to noisy sentences, and AAcc is designed for measuring this ability. 
      Given a bag $\mathcal{B}_i=\{s_{i1},\ldots,s_{in_{s}}\}$ containing both valid and noisy sentences and the attention weights $\{\alpha_{i1},\ldots,\alpha_{in_{s}}\}$ over the sentences, AAcc of this bag is formally defined as follows:
      $$
      \mathrm{AAcc}_{i}=\frac{\sum_{j=1}^{n_{s}} \sum_{k=1}^{n_{s}} \mathbf{I}\left(z_{ij}\right) \mathbf{I}\left(1-z_{ik}\right) \mathbf{I}\left(\alpha_{i j}>\alpha_{i k}\right)}{\sum_{j=1}^{n_{s}} \mathbf{I}\left(z_{ij}\right) \sum_{j=1}^{n_{s}} \mathbf{I}\left(1-z_{ij}\right)}
      $$
      where $\mathbf{I}(\cdot)$ is an indicator function and $z_{ij}$ is a flag indicating whether the sentence $s_{ij}$ is valid. The numerator counts how many valid-noisy sentence pairs show higher attention weight on the valid sentence, and the denominator is the number of valid-noisy sentence pairs is the bag $\mathcal{B}_i$. Then AAcc of the whole test set is computed as $\mathrm{AAcc} = \frac{\sum^{n}_{i=1} \mathrm{AAcc}_i}{n}$, where $n$ is total number of bags.
      
      \item \textbf{F1-Score}.
      We adopt the F1-Score instead of AUC (which is used in \cite{hu2021knowledge}) as the evaluation metric, since the test labels are human-annotated.
      Firstly, we calculate F1-Score on the whole test set as a baseline. 
      Then we construct three test sets with \textit{zero} / \textit{one} / \textit{all} valid sentences in each bag. 
      Finally, we calculate F1-Scores on these three datasets.
      We emphasize that for the case of \textit{zero}, a prediction equaling the bag relation is incorrect, since there are no sentences expressing the relation in the bag.
    \end{itemize}

    \begin{table}[t]
      \centering
      \caption{(\%) AAcc and F1-Scores results on NYT-H, 
      where \textit{Orig.} indicates the original test set, 
      \textit{Zero} indicates the test set in which no sentences are valid,
      \textit{One} indicates the test set in which only one sentence is valid in each bag, 
      \textit{All} indicates the test set in which all sentences are valid.
      }
      \begin{tabular}{lcccccc}
          \toprule
          \multirow{2}*{\textbf{Model}} & \multirow{2}*{\textbf{AAcc}} & \multicolumn{4}{c}{\textbf{F1-Score}} \\
          \cmidrule(lr){3-6}
          ~ & ~  & Orig. & Zero & One & All \\
          \midrule
          PCNN+ATT\cite{lin2016neural} & 52.9 & 60.5 & 63.9 & 72.4 & 75.7 \\
          \midrule
          PCNN+CATT & \textbf{57.3} & \textbf{66} & \textbf{71.1} & \textbf{78.5} & \textbf{80.2} \\
          \bottomrule
      \end{tabular}
      \label{table7}
    \end{table}

    We present the results in Table \ref{table7}, in which CGRE is renamed PCNN+CATT (constraint-aware attention), to highlight the attention mechanisms used.
    From Table \ref{table7}, we observe that: 
    
    (1) PCNN+CATT achieves 4.2\% AAcc improvements over PCNN+ATT, demonstrating that the stronger ability of CATT over ATT to recognize the noisy sentences and to dynamically reduce their attention weights.

    (2) PCNN+CATT achieves 5.5\%, 7.2\%, 6.1\% and 4.5\% F1-Score improvements over PCNN+ATT, in the settings of \textit{Orig.}, \textit{Zero}, \textit{One} and \textit{All}, respectively. Compared with the original selective attention mechanism\cite{lin2016neural}, our designed constraint-aware attention mechanism shows stronger abilities to identify valid sentences in various noisy environments. Specially, all sentences in \textit{Zero} are valid, so it is satisfactory that any sentence obtains higher attention weight. However, PCNN+CATT still achieve a significant advantage in \textit{Zero}. The reason we speculate is that CATT tends to assign higher attention weights to the easily recognized valid sentences.

    \subsection{Ablation Study}
    We perform ablation studies to further evaluate the effects and contributions of different components of CGRE. 
    Note that all ablation experiments are conducted on \textbf{NYT-520K}, for a more accurate evaluation.

    \subsubsection{Effect of Backbone Models}\label{sec3.6.1}
    We evaluate the effects of different backbone models on the performance of CGRE. We experiment with three widely used sentence encoders \textbf{CNN}\cite{zeng2014relation}, \textbf{PCNN}\cite{zeng2015distant} and \textbf{BERT}\cite{kenton2019bert}, and three representative graph encoders \textbf{GCN}\cite{kipf2017semi}, \textbf{GAT}\cite{velivckovic2018graph} and \textbf{SAGE}\cite{hamilton2017inductive}. 
    Compared with GCN, GAT dynamically assigns different attention weights to different neighbors in aggregation computation,
    while SAGE generates embeddings by sampling and aggregating neighboring features instead of training individual embeddings for each node.
    Note that the entity-aware word embedding\cite{li2020self} is only applied in CNN and PCNN, since BERT has integrated word embeddings.

    \begin{table}[t]
      \centering
      \caption{(\%) AUC and macro accuracy of Hits@K of different backbone models on NYT-520K, where \textit{None} indicate the case without the constraint graph (e.g., CNN+\textit{None} is equivalent to CNN+ATT+ENT), playing the baseline role, and $<$100/$<$200 indicates the long-tailed relations with training instances fewer than 100/200.}
      \begin{tabular}{ccccccc}
          \toprule
          \multicolumn{2}{c}{\textbf{Encoder}} & \multicolumn{2}{c}{\textbf{$<$100}} & \multicolumn{2}{c}{\textbf{$<$200}} & \multirow{2}*{\textbf{AUC}}\\
          \cmidrule(lr){1-2}\cmidrule(lr){3-4}\cmidrule(lr){5-6}
          Sent. & Graph & Hits@5 & Hits@10 & Hits@5 & Hits@10 & ~ \\
          \midrule
          \multirow{4}*{CNN} & \textit{None} & 7.5 & 20.0  & 19.2 & 32.4 & 35.0 \\
          \cmidrule(lr){2-7}
          & GCN & 17.5 & 40.0 & 31.2 & 50.0 & 38.2 \\
          & GAT & \textbf{27.5} & \textbf{40.0} & \textbf{39.6} & \textbf{50.0} & \textbf{38.7} \\
          & SAGE & 24.2 & 40.0 & 34.0 & 50.0 & 37.3 \\
          \midrule
          \multirow{4}*{PCNN} & \textit{None} & 15.0 & 35.0 & 28.2 & 45.8 & 35.5 \\
          \cmidrule(lr){2-7}
          & GCN & 19.2 & 40.0 & 32.6 & 50.0 & 41.7 \\
          & GAT & \textbf{23.3} & 40.0 & \textbf{35.2} & 50.0 & \textbf{41.8} \\
          & SAGE & 22.5 & \textbf{65.0} & 34.5 & \textbf{70.8} & 39.3 \\
          \midrule
          \multirow{4}*{BERT} & \textit{None} & 10.0 & 30.0 & 25.0 & 41.7 & 46.1 \\
          \cmidrule(lr){2-7}
          & GCN & \textbf{31.7} & 36.7 & \textbf{43.1} & 47.2 & 46.8 \\
          & GAT & 23.3 & 40.0 & 36.1 & 50.0 & \textbf{48.9} \\
          & SAGE & 25.0 & \textbf{45.0} & 36.6 & \textbf{54.2} & 47.2 \\
          \bottomrule
      \end{tabular}
      \label{table8}
    \end{table}

    We report the experimental results in Table \ref{table8}, from which we observe that:

    (1) compared with the baselines without the constraint graph, any combination achieves significant and consistent improvements on performance of denoising and long-tailed RE. 
    This strongly suggests the effectiveness of the proposed constraint graph;

    (2) among the sentence encoders, BERT shows huge advantages on overall performance against noise and top-5 hit rate on long-tailed relations. 
    However, CNN and PCNN also show competitive performance on the top-10 hit rate;
    
    (3) among the graph encoders, GAT achieves the highest anti-noise performance, independently of the sentence encoder choices. In terms of long-tailed RE, GAT provides the highest improvements for CNN and PCNN, while GCN and SAGE show more benefits for BERT;
    
    (4) the BERT-based CGRE variants achieve significant improvements over the previous Transformer architectural model, DISTRE~\cite{alt2019fine}, which achieves an AUC value of 42.2\% on NYT-520K.

    Although BERT and GAT show significant superiority in this ablation experiment, we still utilize PCNN and GCN as the sentence and graph encoders of vanilla CGRE, for their high efficiency and acceptable accuracy. We retain the choices of other backbone models as the variants of CGRE.

    \begin{figure*}[b]
      \centering
      \subfigure[Performance of GCNs with different layer numbers.]{
        \includegraphics[scale=0.5]{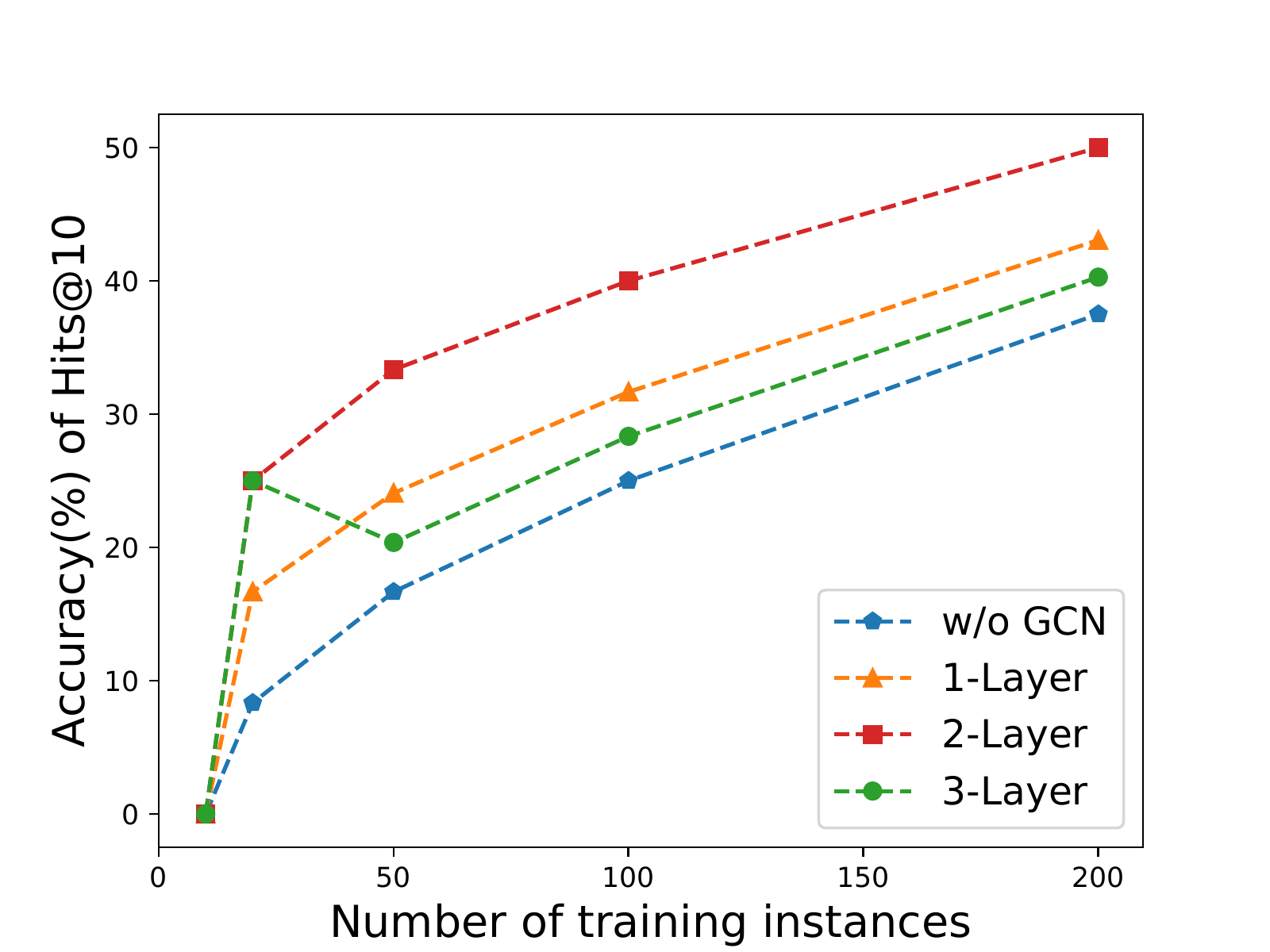}
        \label{figure5a}
      }\quad\quad
      \subfigure[Performance of GCNs with different output options.]{
        \includegraphics[scale=0.5]{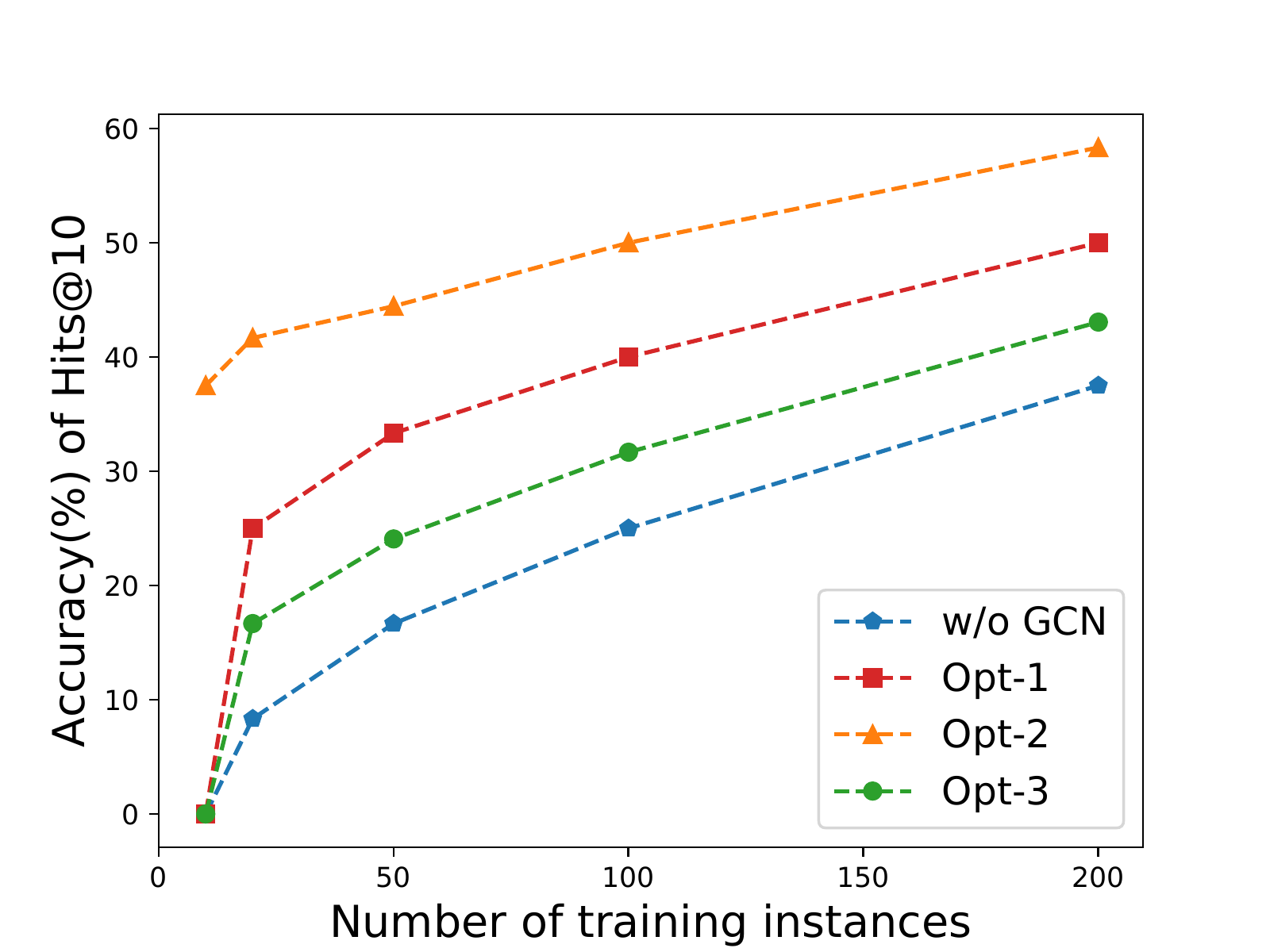}
        \label{figure5b}
      }\quad\quad
      \caption{Performance on the long-tailed relations with different number of training instances.
      }
      \label{figure5}
    \end{figure*}

    \subsubsection{Effect of the Constraint Graph}\label{sec3.6.2}
    As effectiveness of the constraint graph has been confirmed in Sec. \ref{sec3.6.1}, we further explore the effects of different information (i.e., type information and constraint information) in the constraint graph. 
    We briefly denote the \textit{CGRE without the constraint graph} as \textbf{Base}, which is equivalent to PCNN+ATT+ENT. 
    To evaluate the effect of \textit{type information}, we build \textbf{Base+Type}, which concatenates the sentence representations with the corresponding type representations, and then transforms the dimension of the concatenating representation to match that of the relation representation for attention computation.
    To evaluate the effect of \textit{constraint information}, we build \textbf{Base+Const}, which is identical to Base, except that instances violating constraints are removed during training and predicted as NA with 100\% probability during testing.

    We present the results of AUC and macro accuracy of Hits@K in Table \ref{table9}, from which we observe that:

    (1) both the external information of entity types and relation constraints provide significant improvements on denoising and long-tailed RE;

    (2) improvements on long-tailed RE from type information and constraint information are comparable. However, type information shows significant advantages (with 2.4\% AUC improvement over constraint information) on overall performance against noise;

    (3) compared with directly integrating the type information or the hard constraint strategy, our proposed CGRE organically combines these two kinds of information to achieve significant improvements in terms of denoising and long-tailed RE.
    
    \begin{table}[t]
      \centering
      \caption{(\%) AUC and macro accuracy of Hits@K of different variants on NYT-520K.}
      \begin{tabular}{lccccc}
          \toprule
          \multirow{2}*{\textbf{Model}} & \multicolumn{2}{c}{\textbf{$<$100}} & \multicolumn{2}{c}{\textbf{$<$200}} & \multirow{2}*{\textbf{AUC}}\\
          \cmidrule(lr){2-3}\cmidrule(lr){4-5}
          ~ & Hits@5 & Hits@10 & Hits@5 & Hits@10 & ~ \\
          \midrule
          Base & 5.0 & 5.0  & 15.3 & 20.8 & 34.7 \\
          Base+Type & $<$5.0 & 27.8  & 16.2 & 40.9 & 39.4 \\
          Base+Const & 12.5 & 25.0  & 26.2 & 37.5 & 37.0 \\
          \midrule
          CGRE & \textbf{19.2} & \textbf{40.0} & \textbf{32.6} & \textbf{50.0} & \textbf{41.7} \\
          \bottomrule
      \end{tabular}
      \label{table9}
    \end{table}
    
    \subsubsection{Effect of GCNs}
    To investigate the effect of the integration mechanism of GCNs on knowledge transfer between relations,
    we evaluate the long-tailed RE performance of GCNs with different layer numbers and different output options.
    We denote CGRE without GCN as \textbf{w/o GCN}, in which the input embeddings $\mathbf{V}^{(0)}$ are directly output, and denote CGRE with a $k$-layer GCN as 
    \textbf{k-Layer}.
    For a $2$-layer GCN, we design 3 different output options:
    
    \begin{itemize}[leftmargin=*]
      \item \textbf{Opt-1} directly take the representations of last layer, i.e. $\mathbf{V}^{(2)}$, for output;
      \item \textbf{Opt-2} concatenates the representations of last 2 layers, i.e. $[\mathbf{V}^{(1)}; \mathbf{V}^{(2)}]$, for output;
      \item \textbf{Opt-3} concatenates the representations of last 3 layers, i.e. $[\mathbf{V}^{(0)}; \mathbf{V}^{(1)}; \mathbf{V}^{(2)}]$, for output.
    \end{itemize}
    To ensure dimensional matching, the concatenated vectors are transformed by a linear layer before output.
    
    From the results shown in Figure \ref{figure5}, 
    we observe that: 
    
    (1) GCN-based variants consistently outperform CGRE w/o GCN. This confirms the intuition that the neighborhood integration mechanism of GCN could effectively propagate rich information from the head relations to the long-tailed relations.
    
    (2) due to over-smoothing, the increase of number of GCN layers may hurt the performance.
    As shown in Fig. \ref{figure5a}, the 2-layer GCN achieve the best performance in our experiments;
    
    (3) Fig. \ref{figure5b} shows a huge advantage of Opt-2 on long-tailed RE. 
    We speculate the reason is that concatenation with the low-layer integrated representations could enhance the information propagation and feature learning, while concatenation with the non-integrated embeddings may impedes the knowledge transfer among relations.

    \begin{table*}[t]
      \centering
      \caption{Examples for case study. 
      The column \textit{Satisfactory} indicates whether the sentence satisfies the constraint, 
      and column the \textit{Correct} indicates whether the label is correct.
      The words in bold represent the target entities.
      In \textit{Entity Type}, NORP implies nationalities, religious or political groups,
      LANG implies languages, 
      and GPE implies countries, cities or states.
      }
      \begin{tabular}{c|l|c|c|c|c}
          \toprule
          \textbf{ID} & \multicolumn{1}{c|}{\textbf{Sentence}} & \textbf{Entity Type} & \textbf{Satisfactory} & \textbf{Correct} & \textbf{Score} \\
          \midrule
          $s_{1}$ &
          \makecell[l]{With the city so influenced by chinese and \textbf{japanese culture},
          \textbf{asian} \\ cuisine is always an excellent bet.}
          & (NORP, NORP) & Yes & Yes & 0.967 \\
          \midrule
          $s_{2}$ &
          \makecell[l]{In the last several years, the site has housed a french restaurant,
          an \\ \textbf{asian} buffet,
          and as of about three months ago a \textbf{japanese} restaurant...}
          & (NORP, NORP) & Yes & No & 0.009 \\
          \midrule
          $s_{3}$ &
          \makecell[l]{"Three extremes", 125 minutes in \textbf{japanese}, korean, mandarin and \\
          cantonese, this trilogy provides a sampler of short horror films from \\
           high-profile \textbf{asian} directors.}
          & (LANG, NORP) & No & No & 0.003 \\
          \midrule
          \multicolumn{2}{l}{\textit{Relation: /people/ethnicity/included\_in\_group}} & \multicolumn{4}{r}{\textit{Constraint: (LANG, NORP)} } \\
          \midrule
          \midrule
          $s_{4}$ &
          \makecell[l]{Herzen was tireless in his crusade for freeing \textbf{russia} 's serfs and refor-\\ming \textbf{russian} society.}
          & (NORP, GPE) & Yes & Yes & 0.695 \\
          \midrule
          $s_{5}$ &
          \makecell[l]{The \textbf{russian} debate on the treaty is subtly shifting, with new attention \\ to the missiles \textbf{russia}
          will really need.}
          & (NORP, GPE) & Yes & No & 0.032 \\
          \midrule
          $s_{6}$ &
          \makecell[l]{In st.petersburg, \textbf{russia}, summer advantage is giving teenagers early \\ college credit for \textbf{russian} language classes.}
          & (LANG, GPE) & No & No & 0.007 \\
          \midrule
          \multicolumn{2}{l}{\textit{Relation: /people/ethnicity/geographic\_distribution}} & \multicolumn{4}{r}{\textit{Constraint: (NORP, GPE)} } \\
          \bottomrule
      \end{tabular}
      \label{table10}
  \end{table*}

  \subsubsection{Effect of Entity-aware Word Embedding}
  The entity-aware word embedding proposed by \cite{li2020self} shows huge advantages on NYT-570K, and has been applied in several recent works\cite{li2020improving,yu2020tohre}. 
  In this study, we comprehensively evaluate the performance of this module.
  As shown in Fig. \ref{figure4} and Table \ref{table3}, its advantages on NYT-520K and GDS are not as huge as that on NYT-570K, which contains many overlapping entity pairs. This suggests that the overall improvements provided by entity-aware word embedding are mainly due to rote memorization of the entity pairs occurring in the training set. However, the positive effectiveness of the module should not be entirely denied, since it provides 0.8\% and 5.0\% AUC improvements on NYT-520K and GDS, respectively. 
  Hence, we still retain this module in vanilla CGRE for its slight improvements and plug-and-play property.

\begin{figure}[t]
    \centering
    \includegraphics[scale=0.5]{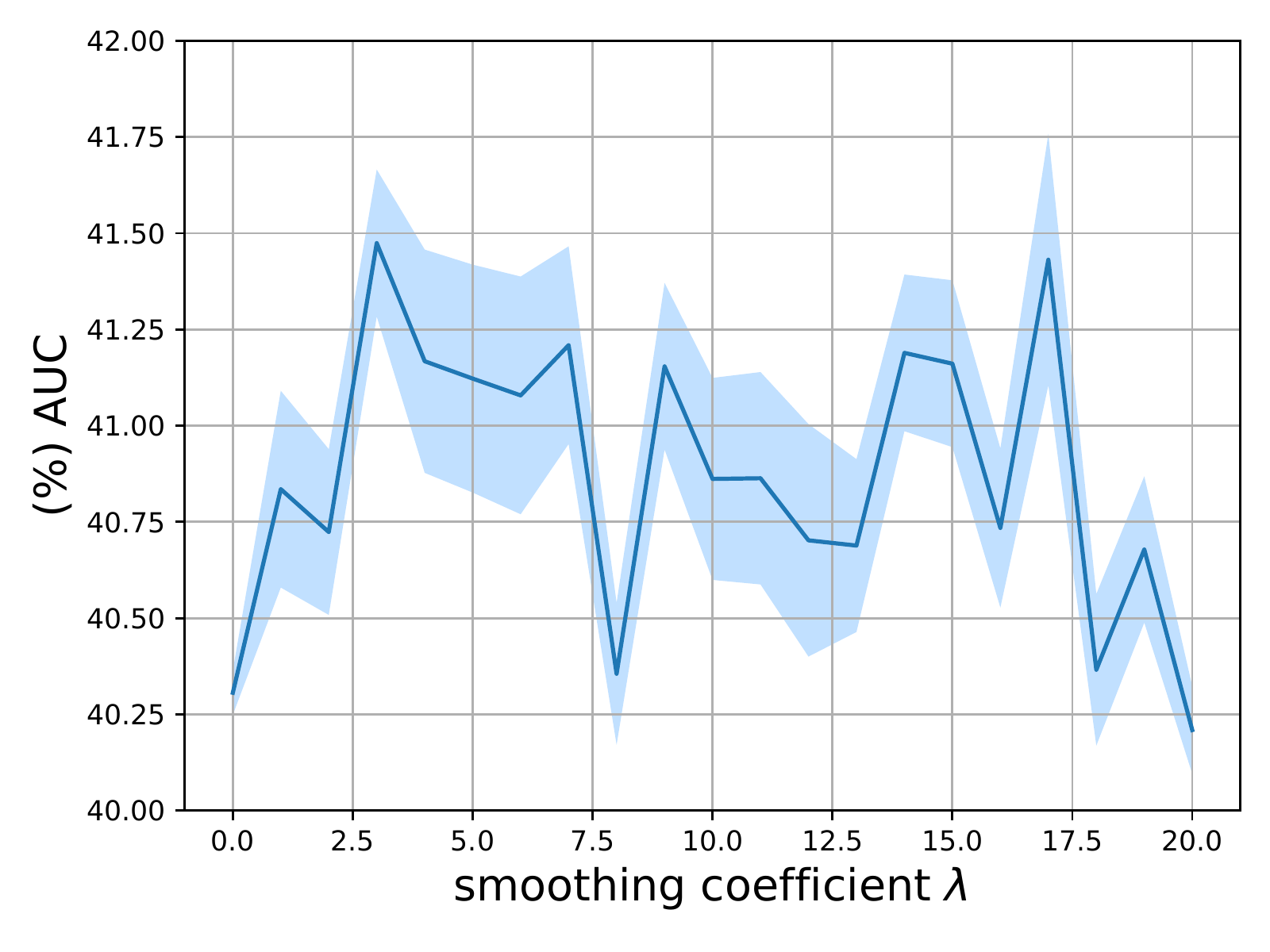}
    \caption{AUC for different smoothing coefficient $\lambda$.}
    \label{figure6}
  \end{figure}

  We further investigate the effect of the smoothing coefficient $\lambda$, which balances the contribution of entity information and word information.
  We experiment with $\lambda \in [0, 20]$, and present the AUC trend in Fig \ref{figure6}. 
  It is clear that the selection of $\lambda$ has a significant impact on the overall performance.
  We observe that the two peaks occur at $\lambda=3$ and $17$, while the trough occurs at $\lambda=8$.

    \subsection{Case Study}
    We present some examples to demonstrate
    how our constraint-aware attention combines the semantic and constraint information.
    As shown in Table \ref{table10}, 
    the incorrect sentences with constraint violation ($s_{3}$ and $s_{6}$) or semantic inconsistency ($s_{2}$ and $s_{5}$)
    are assigned lower scores,
    while the correct sentences ($s_{1}$ and $s_{4}$) score higher.
    With the constraint-aware attention mechanism, CGRE can effectively recognize the valid instances from noisy data.

    \section{Related Work}
    To automatically obtain large scale labeled training data, \cite{mintz2009distant} proposed distant supervision. 
    However, the relation labels collected through distant supervision are not only noisy but also extremely unbalanced.
    Therefore, denoising and long-tailed RE become two major challenges in DSRE.

    \subsection{Denoising in DSRE.}
    Multi-instance learning (MIL), 
    aiming to extract relations of an entity pair from a sentence bag instead of a single sentence,
    is a popular approach to alleviate the noisy labeling problem~\cite{riedel2010modeling,hoffmann2011knowledge,surdeanu2012multi}. 
    Under the MIL framework, various denoising methods are proposed,
    including selective attention mechanism~\cite{lin2016neural,ye2019distant},
    external information integration~\cite{ji2017distant,vashishth2018reside}
    and robust context encoders~\cite{zeng2015distant,alt2019fine}.
    Different from the bag-level prediction of MIL, some studies~\cite{jun2018reinforcement,jia2019arnor} attempt to predict relations at sentence level by selecting trustable instances from training data.

    In this work, we mainly focus on the bag-level prediction. 
    There are two main differences between CGRE and the above MIL methods:
    (1) rather than treat each relation label separately, CGRE attempts to mine the latent dependencies between relations;
    (2) the attention mechanism of CGRE integrates the prior constraint information, 
    which is effective for improving the noise immunity of RE models.

    \subsection{Long-tailed relation extraction in DSRE.}
    \cite{krause2012large} develops an RE system, which can automatically learn the rules for long-tailed relations from Web.
    To generate fewer but more precise rules, \cite{gui2016exploring} applies explanation-based learning to improve the RE system.
    However, the above methods merely handle each relation in isolation, regardless of the implicit associations between relations.
    Therefore, some recent works~\cite{han2018hierarchical, zhang2019long, li2020improving,yu2020tohre} attempt to mine the semantic dependencies between relations from the relation hierarchical tree. 
    To achieve the same objective without the relation hierarchical tree, \cite{gou2020dynamic} proposed DPEN\footnotemark[9], utilizing entity types to implicitly transfer information across different relations.
    \footnotetext[9]{Main differences between DPEN and CGRE are detailed in Sec. \ref{sec:3.4}.}
    Although these works achieve significant improvement for long-tailed relations, they are still far from satisfactory.

    Different from relation hierarchies that mainly contain the hierarchical information between labels,
    the proposed constraint graph involves not only the implicit connection information between labels, but also the explicit prior information of constraints.
    By integrating both kinds of information, our CGRE can effectively handle the noisy and long-tailed labels.

    \subsection{Type constraint information in DSRE.}
    Type information of entities showed great potential on relation extraction~\cite{peng2020learning,zhong2021frustratingly}.
    Some works ~\cite{liu2014exploring, vashishth2018reside} explored the utilization of fine-grained entity type information to improve the robustness of DSRE models.
    However, these methods merely focus on the semantic information of types, 
    i.e., they simply use type information as concatenating features to enrich the sentence semantic representation,
    but neglect the explicit constraint information, which can be effective for recognizing the noisy instances.
    On the contrary, \cite{lei2018cooperative} applies probabilistic soft logic to encode the type constraint information for denoising, but ignores the semantic information of entity types.
    
    As compared with the above methods, CGRE can take full advantage of type information.
    With the constraint-aware attention mechanism,
    CGRE can combine the semantic and constraint information of entity types.
    Moreover, in CGRE, the entity types are used as bridges between the relations, enabling message passing between relation labels.

    \section{Conclusion and Futute Work}
    In this paper, we aim to address both the challenges of label noise and long-tailed relations in DSRE.
    We introduce a novel relation dependency structure, constraint graph, to model the dependencies between relations,
    and then propose a novel relation extraction framework, CGRE, to integrate the information from the constraint graph.
    Experimental results on two popular benchmark datasets have shown that:

    (1) the constraint graph can effectively express the latent semantic dependencies between relation labels;

    (2) our framework can take full advantage of the information in the constraint graph, significantly and consistently outperforming the competitive baselines in terms of denoising and long-tailed RE.

    We believe our exploration for constraint graph shall shed light on future research directions, especially for long-tailed relation extraction.

    Specifically, we plan to explore:
    
    (1) structural extensions of the constraint graph. For example, it would be promising to further improve the performance of long-tailed RE by combining the constraint graph and the relation hierarchical tree;
    
    (2) effective approaches to strengthen the constraint graph. As demonstrated in \cite{zhang2019long} and \cite{christopoulou2021distantly}, KB priors are effective external information for improving DSRE. 
    It is therefore desirable to integrate KB priors into the constraint graph for further performance improvements.

% use section* for acknowledgment
\ifCLASSOPTIONcompsoc
  % The Computer Society usually uses the plural form
  \section*{Acknowledgments}
\else
  % regular IEEE prefers the singular form
  \section*{Acknowledgment}
\fi

This work is supported by the National Key Research and Development Program of China (No. 2016YFC0901902)
and the National Natural Science Foundation of China (No. 61976071, 62071154, 61871020).

% Can use something like this to put references on a page
% by themselves when using endfloat and the captionsoff option.
\ifCLASSOPTIONcaptionsoff
  \newpage
\fi

% trigger a \newpage just before the given reference
% number - used to balance the columns on the last page
% adjust value as needed - may need to be readjusted if
% the document is modified later
%\IEEEtriggeratref{8}
% The "triggered" command can be changed if desired:
%\IEEEtriggercmd{\enlargethispage{-5in}}

% references section

% can use a bibliography generated by BibTeX as a .bbl file
% BibTeX documentation can be easily obtained at:
% http://mirror.ctan.org/biblio/bibtex/contrib/doc/
% The IEEEtran BibTeX style support page is at:
% http://www.michaelshell.org/tex/ieeetran/bibtex/
\bibliographystyle{IEEEtran}
% argument is your BibTeX string definitions and bibliography database(s)
\bibliography{IEEEabrv,ref}
\end{document}